\documentclass[final,12pt]{clear2025} 

\usepackage{makecell}
\usepackage{tikz,pgf}
\usepackage{tikzsymbols}
\usetikzlibrary{positioning, patterns, shapes}
\usepackage{amsmath}
\usepackage{amssymb}
\usepackage{fontawesome}
\usepackage{xr}
\usepackage{algorithm}
\usepackage{algorithmicx}
\usepackage[noend]{algpseudocode}
\usepackage{enumitem}
\algnewcommand{\IfThenElse}[3]{
  \State \algorithmicif\ #1\ \algorithmicthen\ #2\ \algorithmicelse\ #3}

\newcommand\ci{\perp\!\!\!\perp}

\newcommand{\G}{{\mathcal G}}
\newcommand{\F}{{\mathcal F}}
\newcommand{\I}{{\mathcal S}}

\newcommand{\person}{\text{\faChild}}

\DeclareMathOperator{\pa}{pa}

\DeclareMathOperator{\ant}{ant}

\DeclareMathOperator{\mb}{mb}
\DeclareMathOperator{\doo}{do}
\DeclareMathOperator*{\argmax}{argmax}

\DeclareMathOperator{\circright}{\text{\textasteriskcentered{}}\!\!\rightarrow}
\DeclareMathOperator{\circleft}{\leftarrow\!\!\text{\textasteriskcentered{}}}
\DeclareMathOperator{\circcirc}{\text{\textasteriskcentered{}}---\text{\textasteriskcentered{}}}
\DeclareMathOperator{\dyad}{\circ---\circ}

\DeclareMathOperator{\biedge}{\leftrightarrow}

\title[Network Effect Estimation With Contagion And Latent Confounding]{Network Causal Effect Estimation In Graphical Models Of Contagion And Latent Confounding}
\usepackage{times}

\clearauthor{%
 \Name{Yufeng Wu} \Email{sw20@williams.edu}\\
 \addr Williams College
 \AND
 \Name{Rohit Bhattacharya} \Email{rb17@williams.edu}\\
 \addr Williams College
}

\begin{document}

\maketitle

\begin{abstract}%
  A key question in many network studies is whether the observed correlations between units are primarily due to contagion  or latent confounding. Here, we study this question using a segregated graph \citep{shpitser2015segregated} representation of these mechanisms, and examine how uncertainty about the true underlying mechanism impacts downstream computation of network causal effects, particularly under full interference---settings where we only have a single realization of a network and each unit may depend on any other unit in the network. Under certain assumptions about asymptotic growth of the network,  we derive likelihood ratio tests that can be used to identify whether different sets of variables---confounders, treatments, and outcomes---across units exhibit dependence due to contagion or latent confounding. We then propose network causal effect estimation strategies  that provide unbiased and consistent estimates if the dependence mechanisms are either known or correctly inferred using our proposed tests. Together, the proposed methods allow network effect estimation in a wider range of full interference scenarios that have not been considered in prior work. We evaluate the effectiveness of our methods with synthetic data and the validity of our assumptions using real-world networks.
\end{abstract}


\section{Introduction}
\label{sec:intro}

A key question that arises in many network studies is whether observed  correlations between units arise primarily due to \emph{contagion} (peer-to-peer influence) or \emph{latent confounding} (unobserved background factors that lead to similar characteristics for connected individuals) \citep{jackson2013diffusion, rosenbusch2019multilevel}.
This mechanistic knowledge can inform different intervention strategies to influence network-wide changes for certain outcomes of interest.  We study this question in full interference settings---settings where each unit may potentially depend on any other unit in the network, and we only have a single realization of the network---and how this difference can impact downstream computation of network causal effects.  

There is a growing literature on studying interference problems through causal graphical models. \citet{ogburn2014causal} proposed an approach to studying interference problems using causal directed acyclic graphs (causal DAGs). \citet{srinivasan2023graphical} extend this approach to also account for different types of non-ignorable missing data including missingness interference; \citet{ogburn2020causal, bhattacharya2020causal, pena2020unifying} use various interpretations of chain graphs (CGs) to study interference. The aforementioned works focus primarily on the partial interference setting. In the full interference setting, \citet{tchetgen2021auto} propose a chain graph representation for interference along with an estimation strategy, known as \emph{auto-g-computation}, for network effects. However, chain graph models do not admit representations of latent confounding, so the auto-g-computation algorithm cannot be applied in such settings. Full interference work in \citet{ogburn2024causal} and \citet{clark2024causal} have similar limitations. \cite{bhattacharya2024causal} propose a mean-field/message passing algorithms for auto-g-computation. 

Here, we use causal interpretations of segregated graphical models \citep{shpitser2015segregated} to model interference. These models allow for the representation of both contagion and latent confounding. \citet{sherman2018identification} provide a sound and complete identification algorithm for network effects in a certain class of segregated graphs. Their proposed estimation strategy, however, relies on directly applying auto-g-computation and so can only be applied in special cases where latent confounding can be ignored for estimation purposes. Moreover, none of the aforementioned works consider model selection procedures for separating contagion from latent confounding, a question that is of great scientific interest, but relatively understudied in the current literature \citep{shalizi2011homophily, ogburn2018challenges, lee2021network}. We add to the existing literature as follows.
\begin{enumerate}[itemsep=0em]
    \item We propose likelihood ratio tests, based on a coding likelihood \citep{besag1974spatial}, that can be used to distinguish between dependence due to contagion and latent confounding under certain distributional assumptions and assumptions about asymptotic growth of the network.\footnote{{In particular, we require that as the network grows, we also obtain a growing number of units that are at least 6 degrees of separation away from each other, a connection to the small-world principle \citep{watts1998collective}. This may not hold in all network settings, and we examine its viability in Section~\ref{sec:experiments} using five real-world networks.}}
    \item We also propose coding likelihood estimators for network causal effects that are consistent and asymptotically normal under different regimes of dependence due to contagion and latent confounding and operate under weaker network asymptotics than our tests.
\end{enumerate}
Contribution (1) can be seen as extending the existing causal discovery literature in interference by adding new hypothesis tests  for distinguishing between contagion and latent confounding in segregated graphical models under full interference. From a causal inference perspective, contribution (2) extends the auto-g-computation method of \citet{tchetgen2021auto} to handle latent confounding. To our knowledge, neither of these directions have been pursued in prior work. {We envision that (1) and (2) will often be used in conjunction with each other---the tests serve as a check to determine whether auto-g-computation is sufficient, or whether our extension of it is required.} 

\section{Model and Problem Setup}
\label{sec:setup}


We model interference using segregated graphs (SGs) \citep{shpitser2015segregated}, a class of loopless\footnote{Graphs that have no edges of the form $V_i \circcirc V_i$.} mixed graphs where vertices can be connected by directed ($\rightarrow$), bidirected ($\leftrightarrow$), and undirected edges ($-$). A loopless mixed graph $\G(V)$ is  an SG if, (i) Any pair of vertices in $V$ can be connected either by a single edge of any type, or a pair of edges with one being directed and the other being bidirected; (ii) There is no vertex in $V$ that is an endpoint of both a bidirected and undirected edge; (iii) There are no partially directed cycles---a partially directed walk starting and ending at the same vertex $V_i$ consisting of a mix of directed and undirected edges with at least one directed edge.\footnote{As an example, $A - B \rightarrow C - A$ is considered a partially directed cycle, but $A - B - C - A$ is not.} We provide some examples of graphs that do and do not satisfy the SG property in Appendix \ref{app:examples_of_sg}. For brevity, we use $V_i \dyad V_j$ to denote uncertainty about whether $V_i$ and $V_j$ are connected via an undirected or bidirected edge and denote $\G(V)$ as simply $\G$ when the vertex set is clear from context.

\subsection{Statistical models of an SG}
A statistical model of an SG $\G(V)$ is defined as the set of distributions $p(V)$ that \emph{segregated factorize} according to $\G$, or equivalently, satisfy a global Markov property defined in terms of \emph{s-separation} \citep{shpitser2015segregated}, a generalization of the c-separation criterion \citep{studeny1996separation} for chain graphs (segregated graphs with no bidirected edges corresponding to no latent confounding). Briefly, s-separation is similar to c-separation, but generalizes the notion of colliders and collider sections (e.g., $A \rightarrow B \leftarrow C$ and $A \rightarrow B - C \leftarrow D$) to also allow for bidirected edges (e.g., $A\leftrightarrow B \leftarrow C$). A more precise description of s-separation in terms of an augmented graph aka moralization criterion \citep{lauritzen1996graphical} is provided in Appendix~\ref{app:s-separation}. The global Markov property of SGs is as follows: if $X \ci_{\text{s-sep}} Y \mid Z$ in $\G(V)$ then $X \ci Y \mid Z$ in $p(V)$. In our work, to facilitate mechanism discovery, we assume that the converse is also true, which is akin to the \emph{faithfulness} assumption made by many constraint and score-based causal discovery methods \citep{spirtes2000causation}. Thus for statistical models of an SG considered here we have $X \ci Y \mid Z \text{ in } p(V) \iff X \ci_{\text{s-sep}} Y \mid Z \text{ in } \G(V)$.

\subsection{Modeling interference with SGs}

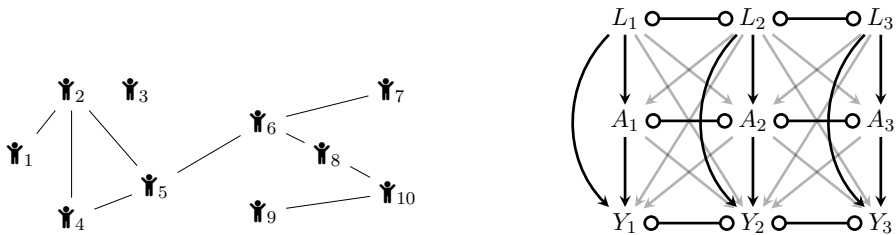
\begin{figure}[t]
    \begin{center}
        \scalebox{0.85}{
    
    \begin{tikzpicture}[>=stealth, node distance=2cm]
        \node (A) at (0, 0) {$\text{\faChild}_4$};
        \node (B) at (0, 2) {$\text{\faChild}_2$};
        \node (C) at (1, 2) {$\text{\faChild}_3$};
        \node (D) at (1.3, 0.5) {$\text{\faChild}_5$};
        \node (E) at (3, 0.1) {$\text{\faChild}_9$};
        \node (F) at (3, 1.5) {$\text{\faChild}_6$};
        \node (G) at (4, 1) {$\text{\faChild}_8$};
        \node (H) at (5, 2) {$\text{\faChild}_7$};
        \node (I) at (5.1, 0.4) {$\text{\faChild}_{10}$};
        \node (J) at (-0.8, 1) {$\text{\faChild}_{1}$};

        \draw (A) -- (B);
        \draw (D) -- (B);
        \draw (A) -- (D);
        \draw (F) -- (H);
        \draw (F) -- (G);
        \draw (E) -- (I);
        \draw (D) -- (F);
        \draw (G) -- (I);
        \draw (J) -- (B);
    \end{tikzpicture}

    \hspace{5em}
    \begin{tikzpicture}[>=stealth, node distance=2cm]
        \tikzstyle{square} = [draw, minimum size=1.0mm, inner sep=3pt]
        \tikzstyle{every node} = [minimum size=1.0mm, inner sep=2pt]

        \begin{scope}[xshift=2cm]
            \path[->, very thick]
            node[] (l1) {$L_1$}
            node[right of=l1] (l2) {$L_2$}
            node[right of=l2] (l3) {$L_3$}
            node[below of=l1, yshift=0.4cm] (a1) {$A_1$}
            node[below of=l2, yshift=0.4cm] (a2) {$A_2$}
            node[below of=l3, yshift=0.4cm] (a3) {$A_3$}
            node[below of=a1, yshift=0.4cm] (y1) {$Y_1$}
            node[below of=a2, yshift=0.4cm] (y2) {$Y_2$}
            node[below of=a3, yshift=0.4cm] (y3) {$Y_3$}

            (l1) edge[] (a1)
            (a1) edge[] (y1)
            (l1) edge[, bend right = 45] (y1)

            (l2) edge[] (a2)
            (a2) edge[] (y2)
            (l2) edge[bend right = 45] (y2)

            (l3) edge[] (a3)
            (a3) edge[] (y3)
            (l3) edge[, bend right = 45] (y3)
            
            (a1) edge[opacity=0.3] (y2)
            (a2) edge[opacity=0.3] (y1)
            (l1) edge[opacity=0.3] (y2)
            (l2) edge[opacity=0.3] (y1)

            (l1) edge[opacity=0.3] (a2)
            (l2) edge[opacity=0.3] (a1)
            (l1) edge[o-o] (l2)
            (a1) edge[o-o] (a2)
            (y1) edge[o-o] (y2)

            (l2) edge[opacity=0.3] (a3)
            (l3) edge[opacity=0.3] (a2)
            (l2) edge[opacity=0.3] (y3)
            (l3) edge[opacity=0.3] (y2)
            (a2) edge[opacity=0.3] (y3)
            (a3) edge[opacity=0.3] (y2)
            (l2) edge[o-o] (l3)
            (a2) edge[o-o] (a3)
            (y2) edge[o-o] (y3)
            ;
        \end{scope}
        
    \end{tikzpicture}
        }
    \vspace{-10pt}
    \end{center}
    \caption{(a) Example of a friendship network $\F$ on 10 units. (b) Partially determined SG pattern $\G_{\dyad}$ on 3 units (lighter edge colors in the figure are only used to improve readability).}
    \label{fig:setup}
\end{figure}

\cite{lauritzen2002chain} provided a causal interpretation for segregated graphs with no bidirected edges, known as chain graphs (CGs). Under this interpretation, a directed edge $V_i \rightarrow V_j$ indicates that $V_i$ is a potential cause of $V_j$ while an undirected edge $V_i - V_j$ represents the possibility of a symmetric causal relationship that reaches an equilibrium state. In the interference literature, this symmetric causal relationship is often interpreted as being the result of a contagious process; see for example \citet{ogburn2020causal, bhattacharya2020causal, tchetgen2021auto}. 

The above interpretations of directed and undirected edges naturally carry forward to causal interpretations of SGs. We now elaborate on the interpretation of bidirected edges.  An SG $\G(V)$ can be thought of as encoding the marginal distribution of a hidden variable chain graph $\G(V \cup H)$ with observed variables $V$ and hidden variables $H$, such that for every edge $V_i \leftrightarrow V_j$ in $\G(V)$, there exists an exogenous hidden variable $H_{ij}$ connecting $V_i$ and $V_j$ in $\G(V \cup H)$ as $V_i \leftarrow H_{ij} \rightarrow V_j$ \citep{shpitser2015segregated}.\footnote{The pattern of hidden variables in the chain graph need not exactly match this description for there to be a bidirected edge. For example, a path $V_i \leftarrow H_1 \rightarrow H_2 \rightarrow V_j$ involving multiple hidden variables is also possible. However, these are all equivalent as they imply the same marginal model, so we opt for the simpler representation.} That is, bidirected edges in an SG represent latent confounding. Equipped with these definitions, we  describe the goals and assumptions of our work.

\subsection{Goals and Graphical Assumptions}

We observe $N$ potentially interconnected units in a network, representing full interference. This network of connections or friendships can be represented by an undirected graph $\F$ with $N$ vertices labeled $\person_1,\ldots, \person_N$ and edges $\person_i - \person_j$ for every pair of units $i, j$ that are connected to each other. For each unit $i$ in the network, we observe a vector of baseline covariates $L_i$, their treatment $A_i$, and their final outcome $Y_i$. That is, our observed data consists of (possibly) dependent realizations $(L, A, Y) = (L_1, A_1, Y_1),\ldots, (L_N, A_N, Y_N)$. In this setting, we set our target of inference to be the expected value of unit-level potential outcomes of the form $\mathbb{E}[Y_i\mid\doo(A=a)] \text{ for } i=1,\ldots,N.$
The quantity $\mathbb{E}[Y_i\mid \doo(a)]$ represents the expected outcome of $\person_i$ when all units in the network receive the \emph{treatment assignment vector} $A=a$. Once these are known, it is easy to compute a variety of other network effects, such as the population average overall effect and spillover effect.

To make this task feasible, we make the following simplifying assumptions: (1) We assume that the friendship network $\F$ is known; (2) The data $(L_1, A_1, Y_1), \ldots, (L_N, A_N, Y_N)$ are drawn from a distribution that is Markov and faithful with respect to an SG $\G$ that is consistent with the \emph{partially determined} SG pattern shown in Figure~\ref{fig:setup}(b) for three units, with the restriction that variables of the same type ($L$, $A$, or $Y$) are always connected by the same type of edge (partially determined refers to the fact that there is uncertainty about whether the $\dyad$ edges are bidirected or undirected). The latter restriction implies, for example, that if the outcome is contagious for two connected units, then it is also contagious for all other connected units (and similarly for latent confounding). These assumptions are similar to those adopted in prior work \citep{bhattacharya2020causal, tchetgen2021auto, ogburn2024causal}, but have the benefit of allowing for certain patterns of latent confounding.

Besides implying different mechanisms, we show in Section~\ref{sec:identification_and_estimation} that, depending on the exact target of inference, the presence of undirected or bidirected edges in different layers\footnote{{We use the term layer to refer to a collection of variables which could be $L$ (the covariate layer), $A$ (the treatment layer), or $Y$ (the outcome layer).}} of the SG may also necessitate the use of different identification and estimation strategies that must be used for computing network causal effects, especially under full interference. 

We end this section by noting an important limitation of our assumptions---by using an SG representation, we apriori rule out the simultaneous existence of contagion and latent confounding in the same layer of $\G$. While this is certainly possible in the real world, smooth globally identified parameterizations of such models have not been proposed yet, and indeed may not be possible in certain cases, as has been shown for linear Gaussian models for graphs that contain ``bows,'' i.e., pairs of vertices such that both $V_i \rightarrow V_j$ and $V_i \leftrightarrow V_j$ exist \citep{drton2009computing, shpitser2015segregated}.

\section{Motivating Example}
Before presenting technical details of our method, we present a hypothetical example inspired by the Networks, Norms and HIV/STI Risk Among Youth (NNAHRAY) study \citep{khan2009incarceration,friedman2008relative}, of how it might be used. Suppose we aim to examine the effect of incarceration on HIV risk. Let $L_i$ represent an individual's past intravenous drug use (note $L_i$ could also be a vector of baseline confounders, but we consider it to be a single variable in this example for simplicity), $A_i$ represent past incarceration status, and $Y_i$ denote HIV status. Connections in the network $\F$ on $N$ individuals in the study population correspond to sexual or drug-use partnership, i.e. $\person_i - \person_j$ exists in $\F$ if the two individuals are such partners. The underlying unknown segregated graph $\G$ is assumed to be one that can be represented by the pattern in Figure~\ref{fig:setup}(b).

A public health policy maker might ask the following questions: (i) Does intravenous drug use in the population spread primarily due to social contagion, or is dependence due to latent factors such as household incomes? (ii) Does incarceration of an individual cause an increased likelihood of incarceration for their partners, or is this dependence also primarily explained by latent factors? Finally, (iii) Is the disease itself (HIV) contagious? The answers to each of these questions can lead to very different policy on containing future infections.

Our tests proposed in Section~\ref{sec:likelihood_ratio_test} can be used to distinguish between such mechanisms and the methods proposed in Section~\ref{sec:identification_and_estimation} can be used to estimate $\mathbb{E}[Y_i\mid \doo(a)]$ and other network effects of interest, such as spillover effects of incarceration of partners. In the example above, our method would confirm background knowledge for (iii) that HIV is in fact a contagious disease that can spread via sexual contact and sharing needles. The answers to (i) and (ii) that pertain more
to social science are less well established, and our test helps answer these from observational data.
Moreover, even if the mechanisms can be fully determined through background knowledge, if latent confounding is present in the $L$ or $Y$ layer, our estimators that extend auto-g-computation are still required for unbiased estimation of the true network effects.

\section{Likelihood Ratio Tests for Determining the SG}
\label{sec:likelihood_ratio_test}

In this section we present likelihood ratio tests that can be used to distinguish between contagion and latent confounding in each layer of a partially determined SG $\G_{\dyad}$ as shown in Figure~\ref{fig:setup}(b). In order to design valid tests for this purpose, the first step is to to find an independence restriction that holds under contagion, but not under latent confounding (or vice versa). However, due to the dependent nature of our data, this alone is not sufficient to design a valid test with standard asymptotic properties. One popular approach to obtaining standard asymptotics with dependent data is to use \emph{coding likelihood} estimators \citep{besag1974spatial, tchetgen2021auto}---estimators where the likelihood factorizes based on selecting conditionally independent samples from the network. We pursue this coding likelihood approach in our work, both for the design of likelihood ratio tests in this section, and for downstream estimation of network effects in the next section. 

In the following, we consider contagion to be our null hypothesis and latent confounding to be the alternative. We follow this convention as contagion is often the assumed mechanism in prior work \citep{ogburn2020causal, tchetgen2021auto, bhattacharya2020causal}. We design three coding likelihood ratio tests to determine the mechanism in each layer $L, A,$ and $Y$ of a partially determined SG $\G_{\dyad}$. Ideally, to prevent propagation of errors, the results of each test should be independent of each other (i.e., the results of one test should not depend on the output of others). In order to design such tests, we first propose conditional independence restrictions for each layer that hold under the null but not the alternative, regardless of the mechanism in other layers. In the following, we will use $\F_i^{(k)}$ as short-hand for all units $\person_j$ that are exactly $k$ degrees of separation away from the unit $\person_i$ in $\F$, i.e., the shortest path from $\person_i$ to $\person_j$ in $\F$ consists of $k$ edges. By convention, $\F_i^{(0)}$ corresponds to $\person_i$ itself. We will also use $X_{\F_i^{(k, k',\ldots)}}$ as shorthand for the $X$ variables of all units $\person_j$ that are $k, k',\ldots$ degrees of separation away from unit $\person_i$. That is,
\begin{align*}
X_{\F_i^{(k, k',\ldots)}} = \{X_j \mid j \in \F_i^{(k)} \cup \F_i^{(k')} \cup \ldots \}.
\end{align*}

\begin{lemma}
\label{lemma:cis}
Given a friendship network $\F$ and a corresponding partially determined SG $\G_{\dyad}$, the following independences hold in each layer under the null hypothesis of contagion but not under the alternative hypothesis of latent confounding, regardless of the mechanism present in other layers. For any unit $i$ in $\F$ for which $\F_i^{(1)}$ and $\F_i^{(2)}$ are not empty, we have,
\begin{align}
    L_i \ci L_{\F_i^{(2)}} &\mid L_{\F_i^{(1)}} \label{eq:l-ci} \\
    A_i \ci A_{\F_i^{(2)}} &\mid A_{\F_i^{(1)}}, L_{\F_i^{(0, 1, 2, 3)}} \label{eq:a-ci} \\
    Y_i \ci Y_{\F_i^{(2)}} &\mid Y_{\F_i^{(1)}}, A_{\F_i^{(0, 1, 2, 3)}}, L_{\F_i^{(0, 1, 2, 3)}}. \label{eq:y-ci}
\end{align}
\end{lemma}
All proofs can be found in the Appendix. Intuitively, the distinct patterns of independence arise due to the presence/absence of colliders depending on the kind of edge that is present in each layer.

As an example, a conditional independence statement that holds under the null in the $A$ layer for $\person_4$ in the friendship network $\F$ in Figure~\ref{fig:setup}(a) and its corresponding SG is,\footnote{A simpler conditional independence  $A_i \ci A_{\F_i^{(2)}} \mid A_{\F_i^{(1)}}, L_{\F_i^{(0, 1)}}$ also holds under the null but not the alternative. However, such independences cannot be used to factorize a coding likelihood, as seen in the proof for Lemma~\ref{lemma:coding}.}
\begin{align*}
A_4 &\ci A_{\F_4^{(2)}} \mid A_{\F_4^{(1)}}, L_{\F_4^{(0, 1, 2, 3)}} 
\implies A_4 \ci \{A_1, A_{6}\} \mid \{A_2, A_5\}, \{L_4, L_1, L_2, L_5, L_6, L_7, L_8\}. 
\end{align*}%

Lemma~\ref{lemma:cis} suggests we can design a likelihood ratio test for each layer by proposing nested models parameterized by a set of parameters $\beta$ and $\gamma$ respectively, where the model parameterized by $\beta$ encodes a set of restrictions in Lemma~\ref{lemma:cis} (depending on the layer being tested) and the model parameterized by $\gamma$ is a strict supermodel that does not. Under full interference, however, proposing such models for the conditional densities of each layer---$p(L), p(A\mid L)$, and $p(Y\mid A, L)$---is infeasible without some dimension reduction assumptions. Here, we will assume parameters for the null and alternative models for each layer do not grow as a function of the size of the network and are shared across all units in the network. For example, in the case of binary treatments in the $A$ layer, we may propose the following nested parametric models,

{\footnotesize
\begin{align}
    p(A_i=1 \mid A_{\F_i^{(1)}}, &L_{\F_i^{(0, 1, 2, 3)}}; \beta) =  \text{expit}(\beta_0 + \beta_1 \sum_{k \in {\F_i^{(1)}}}A_k \ +  \beta_1L_i +\beta_2 \sum_{k \in {\F_i^{(1, 2, 3)}}}L_k)\label{eq:null_a}, \\
    p(A_i=1 \mid A_{\F_i^{(1, 2)}}, &L_{\F_i^{(0, 1, 2, 3)}}; \gamma) =
    \text{expit}(\gamma_0 + \gamma_1 \sum_{k \in {\F_i^{(1)}}}A_k \ +  \gamma_1L_i +\gamma_2 \sum_{k \in {\F_i^{(1, 2, 3)}}}L_k + \gamma_3\sum_{k \in {\F_i^{(2)}}}A_k)\label{eq:alt_a},
\end{align}}%
where the parameters $\beta$ and $\gamma$ are shared across all units $i$ in the network. We note that other parametric forms for these models are possible, e.g., having separate parameters for modeling dependence of $A_i$ on $L_k$ for $k \in \F_i^{(1)}, \F_i^{(2)}$, and $\F_i^{(3)}$ (we use this more flexible model in our experiments), as long as the number of parameters are not a function of the size of the network $N$ and the null and alternatives are nested hypotheses as in \eqref{eq:null_a} and \eqref{eq:alt_a}.

We now show that the parameters can be consistently estimated using a coding likelihood that factorizes based on selecting units that are at least six degrees of separation away from each other in $\F$. We call such a set a $6$-degree separated set and denote it as $\I^6(\F)$, i.e., every distinct $\person_i, \person_j \in \I^6(\F)$ are at least six degrees of separation apart. Later, we will also use $\I^k(\F)$ to denote $k$-degree separated sets.\footnote{As concrete examples, $\{\person_1, \person_9\}$ is an $\I^6(\F)$ set for Figure~\ref{fig:setup}(a), while $\{\person_1, \person_3, \person_4, \person_6, \person_{10}\}$ is an $\I^2(\F)$ set.} Below we present factorizations of coding likelihoods for the null and alternative models and show they hold regardless of the true underlying mechanism---this is important as we want standard asymptotics  for estimation of the null even if the alternative is true and vice versa.

\begin{lemma}
\label{lemma:coding}
Given data $(L, A, Y)$ drawn from a distribution that is Markov and faithful wrt to a partially determined SG $\G_{\dyad}$, the coding likelihood functions for the null and alternative models factorize as follows regardless of the true underlying mechanisms in each layer of $\G_{\dyad}$,

{\small
\begin{minipage}{0.45\textwidth}
\begin{align*}
&{\cal CL}(\beta_L) = \prod_{i \in \I^6(\F)} p(L_i \mid L_{\F_i^{(1)}}; \beta_L)\\
&{\cal CL}(\beta_A) = \prod_{i \in \I^6(\F)} p(A_i \mid A_{\F_i^{(1)}}, L_{\F_i^{(0, 1, 2, 3)}}; \beta_A) \\
&{\cal CL}(\beta_Y) = \prod_{i \in \I^6(\F)} p(Y_i \mid Y_{\F_i^{(1)}}, \{A\cup L\}_{\F_i^{(0, 1, 2, 3)}}; \beta_Y) 
\end{align*}
\end{minipage}
\begin{minipage}{0.45\textwidth}
\begin{align}
&{\cal CL}(\gamma_L) = \prod_{i \in \I^6(\F)} p(L_i \mid \cdot, L_{\F_i^{(2)}}; \gamma_L) \label{eq:coding_l} \\
&{\cal CL}(\gamma_A) = \prod_{i \in \I^6(\F)} p(A_i \mid \cdot, A_{\F_i^{(2)}}; \gamma_A) \label{eq:coding_a} \\
& {\cal CL}(\gamma_Y) = \prod_{i \in \I^6(\F)} p(Y_i \mid \cdot, Y_{\F_i^{(2)}}; \gamma_Y) \label{eq:coding_y}
\end{align}
\end{minipage}
}
\vspace{0.5em}
where $(\cdot,)$ in each case above denotes conditioning on the same set in the null.
\end{lemma}

Intuition for the factorizations is as follows: for any $\person_i$, we condition on a set of variables in $\F_i^{(0, 1, 2, 3)}$. Thus, the rest of the variables appearing in the coding likelihood functions belong to units that are at least 4 degrees of separation away from $\person_i$ (by the antecedent that all units in $\I^6(\F)$ are at least 6 degrees separated from each other). Thus, factorization of, for e.g., ${\cal CL}(\beta_Y)$ is established by showing that $Y_i \ci L_j, A_j, Y_j \mid \cdot$ for all $\person_j$ that are at least 4 degrees of separation from $\person_i$. Having established factorization of the coding likelihood functions, we can now rely on standard maximum likelihood asymptotic theory (similar to \citet{besag1974spatial}) for constructing coding likelihood ratio tests for each layer, as long as the number of units in $\I^6(\F)$ grows with the size of the network. This leads us to a statement about our assumptions on network asymptotics for designing our tests.

\paragraph{Network asymptotics for testing} Let $\G_{\dyad N}$ be a sequence of partially determined SGs with associated friendship networks $\F_N$ for $N=1, 2, \ldots$ such that as $N \to\infty$, the size of at least one 6-degree separated set obtained from $\F_N$ also tends to infinity, i.e. $|\I(\F_N)|\to\infty$.

Under the above network asymptotics and mild regularity conditions stated in the Appendices of \citet{tchetgen2021auto}, the estimates for the unknown parameters of correctly specified null and alternative models obtained by maximizing the coding likelihood functions in Lemma~\ref{lemma:coding} are consistent and asymptotically normal. In Algorithm~\ref{alg:lr-test} we then propose our final coding likelihood ratio tests for determining the mechanism in each layer of a partially determined SG $\G_{\dyad}$.

Based on Lemma~\ref{lemma:cis} and Lemma~\ref{lemma:coding} and the consistency of the coding likelihood ratio estimates for the parameters of correctly specified null and alternative models, the test for each layer possess the desired asymptotic properties: type-I error control at a specified significance level $\alpha$ and power tending to $1$ as $N \to\infty$. This is also confirmed empirically with our experiments in Section~\ref{sec:experiments}.

Note that many possible $\I^6(\F)$ sets of potentially different sizes can be obtained from a single network $\F$. Ideally, we would want to find the largest such set for estimating our parameters. However, finding the maximum sized $k$-degree separated set is well-known to be NP-hard \citep{robson1986algorithms, myrvold2013fast, eto2014distance}. Here, we opt for a greedy approach to finding maximal $k$-degree separated sets and find that this works well in practice. We also note a well-known tradeoff of coding likelihood estimators is that asymptotic guarantees come at the cost of sample efficiency \citep{besag1974spatial}. The network asymptotics for our tests are particularly interesting in this regard---the small-world principle is a popular hypothesis that in real-world networks every pair of units is no more than $6$ degrees of separation apart \citep{watts1998collective}. Fortunately, our tests make use of units in $\I^6(\F)$ that exactly meet this upper bound, presenting an interesting connection between the connectedness of networks and the testability of mechanisms.

\begin{algorithm}[t] 
\caption{for determining edge types in $\G_{\dyad}$}
\label{alg:lr-test}
\begin{algorithmic}[1]
\State \textbf{Inputs:} Partially determined SG $\G_{\dyad}$; Network $\F$; \ Data $L, A, Y$;\ Significance level $\alpha$.
\State Obtain a maximal $\I^6(\F)$ such that for each $\person_i$ in the set, $\F_i^{(1)}$ and $\F_i^{(2)}$ are not empty
\For{each type of variable $X$ in $``L", ``A", ``Y"$}
\vspace{0.5em}
\State $\widehat{\beta}_X \gets \argmax_{\beta_X} {\cal CL}(\beta_X)$  and  $\widehat{\gamma}_X \gets \argmax_{\gamma_X} {\cal CL}(\gamma_X)$
\vspace{0.25em}
\State $\Lambda \gets -2(\log {\cal CL}(\widehat{\beta}_X) - \log{\cal CL}(\widehat{\gamma}_X))$ and $k \gets |\beta_X| - |\gamma_X|$
\vspace{0.5em}
\If{$P\left( \chi^2_k \geq \Lambda \right) < \alpha)$} \  replace each $X_i \dyad X_j$ in $\G_{\dyad}$ with $X_i \leftrightarrow X_j$
\Else \ \  replace each $X_i \dyad X_j$ in $\G_{\dyad}$ with $X_i - X_j$
\EndIf
\EndFor
\State \textbf{return} the now fully determined $\G_{\dyad}$
\end{algorithmic}
\end{algorithm}

\section{Identification and Estimation of Network Effects}
\label{sec:identification_and_estimation}

The identification and estimation methods presented in this section assume a known SG $\G$, or one that is correctly inferred using Algorithm~\ref{alg:lr-test}. We start with an identification result for the target.

\begin{theorem}
    \label{theorem:id_formula}
    Given a hidden variable CG $\G(V \cup H)$ whose observed margin can be represented by any SG $\G(V)$ that is compatible with the assumptions of Figure~\ref{fig:setup}(b), the target parameter is identified as, $\mathbb{E}[Y_i\mid \doo(A=a)]=\sum_{L, Y} p(L) \times p(Y \mid A=a, L) \times Y_i.$
\end{theorem}

That is, network ignorability \citep{tchetgen2021auto} is satisfied regardless of mechanisms present at each layer. While the identifying functional for our target parameter does not depend on mechanisms, we see in Section~\ref{subsec:estimation} that mechanistic knowledge, particularly in the $L$ and $Y$ layer, is still essential for estimation purposes. For completeness, we also provide an example of a different target parameter where the identifying functional can change depending on mechanisms.

Consider the unit level potential outcome where we intervene on the outcomes of all other units except $\person_i$, i.e., $\mathbb{E}[Y_i \mid \doo(Y_{-i} = y_{-i})]$. 
When contagion is present in the $Y$ layer, the identifying functional for this query is, $\sum_{L, A, Y_i}p(L, A)\times p(Y_i\mid Y_{-i}=y_{-i}, A, L) \times Y_i$; when latent confounding is present in the $Y$ layer, the identifiying functional is simply $\sum_{Y_i}p(Y_i)\times Y_i$. These functionals align with the interpretation of undirected and bidirected edges---interventions on $Y_{-i}$ may have an effect on $Y_i$ under contagion but not latent confounding. The proofs are in Appendix~\ref{app:proof_lemma_different_id}.


\subsection{Extending auto-g-computation to account for latent confounding}
\label{subsec:estimation}

\begin{algorithm}[t]
\caption{for estimating $\mathbb{E}[Y_i \mid \text{do}(A = a)]$ for every $\person_i$ in the network}
\label{alg:network-effects}
\begin{algorithmic}[1]
\State \textbf{Inputs:} SG $\G$; Friendship network ${\cal F}$; Data $L, A, Y$; Treatment assignment vector $a$.
\State Obtain a maximal ${\I}^2(\F)$ set
\If {each $L_i \dyad L_j$ in $\G$ is undirected}
\State $\widehat{\theta}_{L_i - L_j} \gets \argmax_{\theta_{L_i - L_j}} {\cal CL}(\theta_{L_i - L_j}) $ and $L_{\text{draws}} \gets \text{Gibbs sampler $L$}\big(\G; \widehat{\theta}_{L_i-L_j})\big)$ \label{alg:l-layer-undir}
\Else
\State Obtain a maximal $\widetilde{\I}^2(\F)$ set of isomorphic local structures
\State $\widehat{\theta}_{L_i \leftrightarrow L_j} \gets \argmax_{\theta_{L_i \leftrightarrow L_j}} {\cal CL}(\theta_{L_i \leftrightarrow L_j})$ and $L_{\text{draws}} \gets \text{ draws from } p(L; \widehat{\theta}_{L_i \leftrightarrow L_j}) $ \label{alg:l-layer-bidir}
\EndIf
\If {each $Y_i \dyad Y_j$ in $\G$ is undirected}
\State $\widehat{\theta}_{Y_i - Y_j} \gets \argmax_{\theta_{Y_i - Y_j}} {\cal CL}(\theta_{Y_i - Y_j})$ and $Y_{\text{draws}} \gets \text{Gibbs sampler } Y\big(\G; \widehat{\theta}_{Y_i-Y_j}; L_{\text{draws}}, a\big)$ \label{alg:y-layer-undir}
\Else {}

\State Fit an outcome regression model $\mathbb{E}[Y_i\mid A_{\F_i^{(0, 1)}}, L_{\F_i^{(0, 1)}}; \theta_{Y_i \leftrightarrow Y_j}]$ using ${\cal CL}(\theta_{Y_i \leftrightarrow Y_j})$ \label{alg:y-layer-bidir-1}

\State // Below $L^{(m)}$ denotes the $m^{\text{th}}$ row of $L_{\text{draws}}$
\State $Y_{\text{draws}} \gets$ $M\times N$ matrix where entry $m, i = \mathbb{E}[Y_i \mid a, L^{(m)}; \widehat{\theta}_{Y_i \leftrightarrow Y_j}]$ \label{alg:y-layer-bidir-2}
\EndIf
\State \Return $\widehat{\mathbb{E}}[Y_i | \text{do}(a)], \ldots, \widehat{\mathbb{E}}[Y_N | \text{do}(a)] \gets \text{empirical averages of outcomes for each $\person_i$ in } Y_\text{draws}$
\end{algorithmic}
\end{algorithm}

We now present a new method for network effect estimation that extends the auto-g-computation method of \citet{tchetgen2021auto} to  account for latent confounding. We first provide a brief description of auto-g-computation before describing our modifications.  We return to our target of inference, identified in Theorem~\ref{theorem:id_formula}, which depends on densities $p(L)$ and $p(Y\mid A, L)$. 

Auto-g-computation overcomes the challenge of having access to only a single realization of the network under full interference by drawing a number of independent samples of $L$ and $Y$  from the densities $p(L)$ and $p(Y\mid A, L)$.  The method proposes a Gibbs sampler that draws independent realizations of $L$ and $Y$ from these densities given access to just univariate conditionals $p(L_i \mid L_{-i})$ and $p(Y_i \mid Y_{-i}, A, L)$ for each $\person_i$ in the network. To  make estimation of these conditionals feasible from a single realization, \citet{tchetgen2021auto} assumes that the $L$ and $Y$ layer are connected by undirected edges, which simplifies these conditionals to depend only on adjacent units in the network, i.e. $p(L_i
 \mid L_{\F_i^{(1)}})$ and $p(Y_i \mid Y_{\F_i^{(1)}}, A_{\F_i^{(0, 1)}}, L_{\F_i^{(0, 1)}})$ respectively. Finally, similar to our assumptions in Section~\ref{sec:likelihood_ratio_test}, auto-g-computation assumes that the parameters $\theta_{L_i - L_j}$ and $\theta_{Y_i - Y_j}$ used to parameterize  these univariate conditional distributions are shared across the network.

Thus, to estimate $\mathbb{E}[Y_i \mid \doo(a)]$ for each $\person_i$, auto-g-computation proceeds by (i) Estimating parameters of the univariate conditionals via maximizing coding likelihoods; (ii) Using Gibbs sampling to draw several, say $M$, independent realizations of $L$ and $Y$ from $p(L)$ and $p(Y\mid A=a, L)$, where each realization from the sampler is an $N\times 1$ vector corresponding to values of $L$ and $Y$ for every unit in the network under a treatment assignment $A=a$; and (iii) Taking the empirical average of each $Y_i$ from these $M$ draws as the final estimate for each $\mathbb{E}[Y_i \mid \doo(a)]$. 

While auto-g works well when edges among $L$ and $Y$ are undirected, it leads to biased estimates if either layer contains bidirected edges. This is because the simplification of the univariate conditionals used for Gibbs sampling no longer holds due to conditioning on colliders. By a simple s-separation argument, $p(L_i \mid L_{-i}) \not= p(L_i \mid L_{\F^{(1)}})$ when the $L$ layer contains bidirected edges and $p(Y_i \mid Y_{-i}, A, L) \not= p(Y_i \mid Y_{\F_i^{(1)}}, A_{\F_i^{(0, 1)}}, L_{\F_i^{(0, 1)}})$ when the the $Y$ layer contains bidirected edges. We propose modifications to auto-g-computation to account for these cases. Our full procedure  is described in Algorithm~\ref{alg:network-effects}; the Gibbs samplers are described in Appendix~\ref{app:gibbs_samplers}. If the else statements in our algorithm do not execute, then there are undirected edges in both the $L$ and $Y$ layer, and our procedure reduces to the original auto-g-computation method (${\cal CL}(\theta_{L_i - L_j})$ and ${\cal CL}(\theta_{Y_i - Y_j})$ in lines~\ref{alg:l-layer-undir} and~\ref{alg:y-layer-undir} are coding likelihood functions used to compute parameter estimates when the edges in both layers are undirected). Thus, we focus on explaining the else statements in Algorithm~\ref{alg:network-effects}.

Define a \emph{local structure} $S$ as a frequently occurring isomorphic subgraph of $\F$ (i.e., a subgraph with the same structure if the labels of the nodes are ignored). Examples of local structures include dyads and triangles (cliques consisting of $3$ units, e.g., $\person_2, \person_4, \person_5$ in Figure~\ref{fig:setup}(a)). For a chosen granularity of local structures, we then propose a parameterization of local structures $p(L_S; \theta_{L_i \leftrightarrow L_j})$ that is shared across the network. For example, under a multivariate normal assumption on $p(L)$ where we parameterize dyad local structures, we would propose $\theta_{L_i \leftrightarrow L_j}$ to be the parameter vector $(\mu_i, \sigma_{ii}, \sigma_{ij})$ corresponding to the mean, variance, and covariance shared across all units (this encodes marginal independences between non-adjacent units, as the covariance between non-adjacent units is $0$ in the model definition).
In the case of discrete data, one could also use log-linear parameterizations described in \citet{rudas2010marginal, forcina2010marginal, evans2013marginal}.\footnote{As the size of the local structure increases, so does the size of the parameter vector, corresponding to more flexible modeling assumptions. In our experiments, we use the simplest possible local structure of dyads, however, we only require that the size of the parameter vector does not grow as a function of the size of the network.} 

Consistently estimating the parameters $p(L_S; \theta_{L_i \leftrightarrow L_j})$ allows us to draw independent realizations of $L$ from a distribution $p(L)$ that is Markov wrt $\G_L(V)$ when the $L$ layer has bidirected edges. This is done in line~\ref{alg:l-layer-bidir} of Algorithm~\ref{alg:network-effects} after obtaining maximum coding likelihood estimates of the parameters. The coding likelihood ${\cal CL}(\theta_{L_i \leftrightarrow L_j})$ in line~\ref{alg:l-layer-bidir} is defined over a collection of local structures $\widetilde{\I}^{2}(\F)$ such that for any $S, S' \in \widetilde{\I}^{2}(\F)$ and for every pair of units, $\person_i \in S$ and $\person_j \in S'$,  $\person_i$ and $\person_j$ are at least 2 degrees of separation away from each other.\footnote{An example of $\widetilde{\I}^{2}(\F)$ in Figure~\ref{fig:setup}(a) when considering dyads as local structures is $\{ \person_1 - \person_2, \person_6 - \person_7, \person_9 - \person_{10} \}$.}

After line~\ref{alg:l-layer-bidir} of Algorithm~\ref{alg:network-effects}, we have $M$ independent draws of $L$ from either an undirected or bidirected model of $p(L)$. The remaining steps do not depend on which one we drew from. We now focus on how the algorithm finishes effect estimation when the $Y$ layer has bidirected edges. Notice that the functional in Theorem~\ref{theorem:id_formula} can be further simplified to $\sum_L p(L) \times \mathbb{E}[Y_i \mid A=a, L]$. Given $M$ independent draws of $L$, the only task that remains then is to estimate the outcome regression $\mathbb{E}[Y_i \mid A, L]$ from which we can get $M$ predictions $Y_i$ given each draw of $L$ and $A=a$ and average these predictions to obtain an estimate of $\mathbb{E}[Y_i \mid \doo(a)]$. When the $Y$ layer contains bidirected edges, we can easily obtain maximum coding likelihood estimates for this outcome regression model using a function ${\cal CL}(\theta_{Y_i \leftrightarrow Y_j})$ defined on units in ${\cal S}^2(\F)$. The estimation and prediction steps are executed in lines~\ref{alg:y-layer-bidir-1} and~\ref{alg:y-layer-bidir-2}. For completeness, we present the factorizations of all coding likelihood functions in Algorithm~\ref{alg:network-effects} in the following Lemma. The difference in factorizations below emphasize the need to know the underlying mechanisms in the $L$ and $Y$ layer, but not necessarily the $A$ layer.

\begin{lemma}
\label{lemma:coding_ricf}
Given data $(L, A, Y)$ drawn from a distribution that is Markov wrt to an SG $\G$ that is compatible with the assumptions in Figure~\ref{fig:setup}(b), the coding likelihood function for modeling $p(L)$ and $p(Y \mid A, L)$ factorize depending on the edges present in the $L$ and $Y$ layers as,


{\footnotesize
\begin{minipage}{0.45\textwidth}
\begin{align*}
    \hspace{-2em}&{\cal CL}(\theta_{L_i - L_j}) = \prod_{i \in \I^2(\F)} p(L_i \mid L_{\F_i^{(1)}}; \theta_{L_i - L_j}) \\
    \hspace{-2em}&{\cal CL}(\theta_{L_i \leftrightarrow L_j}) = \prod_{S \in \widetilde{\I}^2(\F)} p(L_S; \theta_{L_i \leftrightarrow L_j})
\end{align*}
\end{minipage}
\begin{minipage}{0.45\textwidth}
\begin{align*}
    \hspace{-2em}&{\cal CL}(\theta_{Y_i - Y_j}) = \prod_{i \in \I^2(\F)} p(Y_i \mid Y_{\F_i^{(1)}}, \{A \cup L\}_{\F_i{(0, 1)}}; \theta_{Y_i - Y_j}) \\
    \hspace{-2em}&{\cal CL}(\theta_{Y_i \leftrightarrow Y_j}) = \prod_{i \in \I^2(\F)} p(Y_i \mid \{A \cup L\}_{\F_i{(0, 1)}}; \theta_{Y_i \leftrightarrow Y_j})
\end{align*}
\end{minipage}
}

\end{lemma}

Maximizing the above functions gives us consistent and asymptotically normal estimates of the corresponding parameters, and thus the target parameters through Algorithm~\ref{alg:network-effects}, as long as we have the following asymptotic growth of the network (which could be considered weaker than the one proposed for our tests, since it only relies on 2-degree separated sets and local structures).

\paragraph{Network asymptotics for estimation} Let $\G_N$ be a sequence of SGs compatible with  Figure~\ref{fig:setup}(b) and with associated friendship networks $\F_N$ for $N=1, 2, \ldots \infty$ such that as $N \to\infty$, the sizes of at least one 2-degree separated set of individuals and one 2-degree separated set of isomorphic local structures obtained from $\F_N$ also tend to infinity,  i.e. $|\I^2(\F_N)|\to\infty$ and $|\widetilde{\I}^2(\F_N)|\to\infty$.


\section{Experiments, Discussion, and Conclusions}
\label{sec:experiments}

We performed experiments to evaluate the correctness of the proposed methods using both synthetic data and semi-synthetic data with known ground truths\footnote{While it would be ideal to evaluate our methods with real-world data, accompanying ground-truths are typically not available in the absence of a gold standard randomized trial (see \citet{keith2023rct} for discussion). Finding publicly available network trial data, however, is challenging, and we were unable to find a suitable one for our experiments.}. Additionally, we tested the validity of our key network assumption---that sufficiently large sets $\I^6(\F)$ can be found---using various real-world social networks. Code for our experiments is available in this  \href{https://github.com/yufeng-wu/interference_segregated_graph}{GitHub repository}.

\subsection{Evaluation of Likelihood Ratio Tests}

We first evaluate our likelihood ratio tests in Algorithm~\ref{alg:lr-test} using effective sample sizes of $200, 500$, $1000, 2000, 3000$, $ 4000, \text{ and } 5000$. These numbers represent sizes of a maximal $\I^6(\F)$ obtained from a synthetic network of $200,000$ units, where each unit has $1$ to $6$ randomly assigned neighbors. We create two configurations of SGs compatible with the partially determined SG in Figure~\ref{fig:setup}(b), one with undirected edges in all layers and another with bidirected edges in all layers; details of the underlying data generating processes are in Appendix~\ref{app:experiment-details}.

\begin{figure*}[t]
\centering
\includegraphics[width=0.95\linewidth]{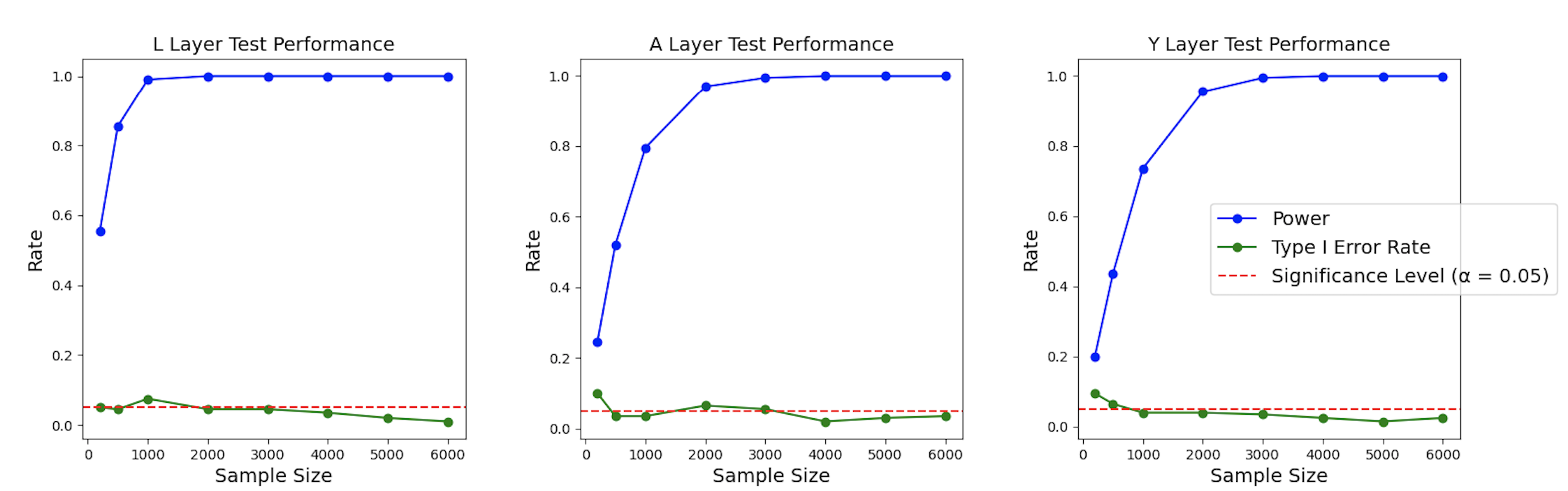}
\caption{Performance of coding likelihood ratio tests with varying sample sizes.}
\label{fig:test_performance_plots}
\end{figure*}

We run $200$ trials at each effective sample size to calculate power and type-I error rate. Figure~\ref{fig:test_performance_plots} shows the results. The results show that our tests are well calibrated, with power approaching $1$ quite quickly as the size of $\I^6(\F)$ increases and type-I error controlled at the desired level of $\alpha=0.05$.

Since real-world networks can be smaller and denser, we also evaluate whether sufficiently large $\I^6(\F)$ can be retrieved in practice by analyzing 10 real networks from the Stanford Large Network Dataset Collection (SNAP) \citep{snapnets}. In each network, we identify 100 maximal 6-degree separated sets, discard units for whom $\F_i{^{(1)}}$ and $\F_i{^{(2)}}$ are empty, as our tests rely on data from these neighborhoods, and retain the largest maximal set, denoted as $|\I^6(\F)|^*$.  Table~\ref{table:5-hop} shows that node usage varies significantly with network density. In denser networks like Twitch Gamers, we could use only $0.02\%$ of units as effective samples for our tests, but in sparser networks such as Deezer RO, we could use $2.85\%$ of the units. These findings indicate that our tests perform best in networks that are both large and relatively sparse. There is also some evidence that virtual social networks are more dense than non-virtual ones \citep{bakhshandeh2011degrees, archiveSysomosDegrees, facebookThreeHalf} due to the ease of forming connections on the internet, so our method may have more samples in non-virtual settings (such as our motivating example).

To evaluate the effectiveness of our likelihood ratio tests on a variety of network topologies, we also conducted semi-synthetic experiments using five real-world networks from Table~\ref{table:5-hop}. For each network, we generate synthetic data on $L$, $A$, and $Y$ using the same rules as in the synthetic version, and we run Algorithm~\ref{alg:lr-test} 500 times. The results are summarized in Table~\ref{tab:semi-synthetic}.

We make the following observations: 1) Type I error rate is well controlled at the desired significance level ($\alpha = 0.05$) across all networks; 2) For a given layer, the power of our test is generally higher for networks where we are able to obtain larger effective samples; and 3) For a given network, the power of our test is generally the highest at the $L$ layer. Distinguishing latent confounding from contagion is slightly more challenging for the $A$ layer and the most challenging for the $Y$ layer.

These results are consistent with the synthetic experiment results in Figure~\ref{fig:test_performance_plots}, showing that our likelihood ratio tests work well on a variety of network topologies. Our synthetic experiments generate networks where each unit has 1 to 6 randomly assigned neighbors. In real-world networks, however, the maximal degree and average degree are much higher. For example, in Deezer HR, the maximum and average degrees are 420 and 18.26; in LastFM Asia, these numbers are 216 and 7.29.

\begin{table*}[]
\centering
\resizebox{0.9\textwidth}{!}{
\begin{tabular}{lrrrr}
\hline
\textbf{Dataset} & \textbf{\makecell{\# of Nodes}} & \textbf{\makecell{\# of Edges}} & \textbf{\makecell{$|\I^6(\F)|^*$}} & \textbf{\makecell{Node Usage}} \\
\hline
Social Circles: FB \citep{leskovec2012learning}    & 4,039   & 88,234    & 3     & 0.07\% \\
GitHub Social Network \citep{rozemberczki2019multiscale}       & 37,700  & 289,003   & 211   & 0.56\% \\
Deezer Europe \citep{feather} & 28,281  & 92,752    & 476   & 1.68\% \\
Deezer HR \citep{rozemberczki2019gemsec}            & 54,573  & 498,202   & 327   & 0.60\% \\
Deezer HU \citep{rozemberczki2019gemsec}            & 47,538  & 222,887   & 488   & 1.03\% \\
Deezer RO \citep{rozemberczki2019gemsec}            & 41,773  & 125,826   & 1,192 & 2.85\% \\
GEMSEC FB Artists \citep{rozemberczki2019gemsec}      & 50,515  & 819,306   & 206   & 0.41\% \\
GEMSEC FB Atheletes \citep{rozemberczki2019gemsec}    & 13,866  & 86,858    & 96    & 0.69\% \\
LastFM Asia \citep{feather} & 7,624   & 27,806    & 159   & 2.09\% \\
Twitch Gamers \citep{rozemberczki2021twitch} & 168,114 & 6,797,557 & 29    & 0.02\%
\end{tabular}
}
\caption{Maximal $\I^6(\F)$ sets in SNAP networks. Node Usage $= {|\I^6(\F)|}^* / \text{\# of Nodes}$.}
\label{table:5-hop}
\end{table*}

\begin{table*}[]
\centering
\resizebox{0.9\textwidth}{!}{
    \small
    \renewcommand{\arraystretch}{1.2}
    \begin{tabular}{lc | cc | cc | cc}
        \hline
        \textbf{} & \textbf{} & \multicolumn{2}{c|}{\textbf{$L$ Layer}} & \multicolumn{2}{c|}{\textbf{$A$ Layer}} & \multicolumn{2}{c}{\textbf{$Y$ Layer}} \\
        \textbf{Network}
        & \textbf{Average $|\I^6(\F)|^*$} & \textbf{Type I} & \textbf{Power} & \textbf{Type I} & \textbf{Power} & \textbf{Type I} & \textbf{Power} \\
        \hline
        LastFM Asia \citep{feather}  & 139.4  & 0.056  & 0.416   & 0.058  & 0.198   & 0.038  & 0.124  \\
        Deezer HR  \citep{rozemberczki2019gemsec}  & 297.7  & 0.046  & 0.490   & 0.028  & 0.358  & 0.052  & 0.044  \\
        Deezer Europe \citep{feather} & 451.1 & 0.044  & 0.392  & 0.046  & 0.324  & 0.056  & 0.416  \\
        Deezer HU \citep{rozemberczki2019gemsec}   & 464.6  & 0.058  & 0.878  & 0.058  & 0.398  & 0.058  & 0.652  \\
        Deezer RO  \citep{rozemberczki2019gemsec}  & 1147.0 & 0.050  & 0.998  & 0.030  & 0.916  & 0.038  & 0.872  \\
    \end{tabular}
    }
    \caption{Performance of Likelihood Ratio Tests on semi-synthetic data. Average $|\I^6(\F)|^*$ is simply the mean of effective sample sizes across. Type I represents type I error rate.}
    \label{tab:semi-synthetic}
\end{table*}

\subsection{Evaluation of Causal Effect Estimation Method}

There are a total of eight 
possible SGs compatible with the partially determined SG $\G_{\dyad}$ shown in Figure \ref{fig:setup}(b). For brevity, we denote these SGs using a three-letter code, where each letter corresponds to a layer and is either `U' (for undirected) or `B' (for bidirected). For example, BUB represents the case where the $L$ layer is connected by $\leftrightarrow$, the $A$ layer by $-$, and the $Y$ layer by $\leftrightarrow$.

We do not evaluate the UUU case because this is already studied in prior work. For each of the other cases, we repeat the following 100 times: 1) Randomly generate friendship networks of different sizes ($500, 1000$, $2000, 3000, 4000$, and $5000$) where units have an average of $5$ and at most $10$ neighbors, and 2) Using data generated from these networks, estimate the population average overall effect as the contrast
$
\mathbb{E}[Y \mid \text{do}(\textbf{1})] - \mathbb{E}[Y \mid \text{do}(\textbf{0})] = \frac{1}{N} \sum_{i=1}^{N} \bigl(\mathbb{E}[Y_i \mid \text{do}(\textbf{1})] - \mathbb{E}[Y_i \mid \text{do}(\textbf{0})]\bigl),
$
where bold-faced $\textbf{1}$ indicates a vector of treatment assignments where everyone in the network receives $A_i = 1$, and similarly for $\textbf{0}$, using both Algorithm~\ref{alg:network-effects} and standard auto-g-computation.

\begin{figure*}[t]
\centering
\includegraphics[width=0.85\linewidth]{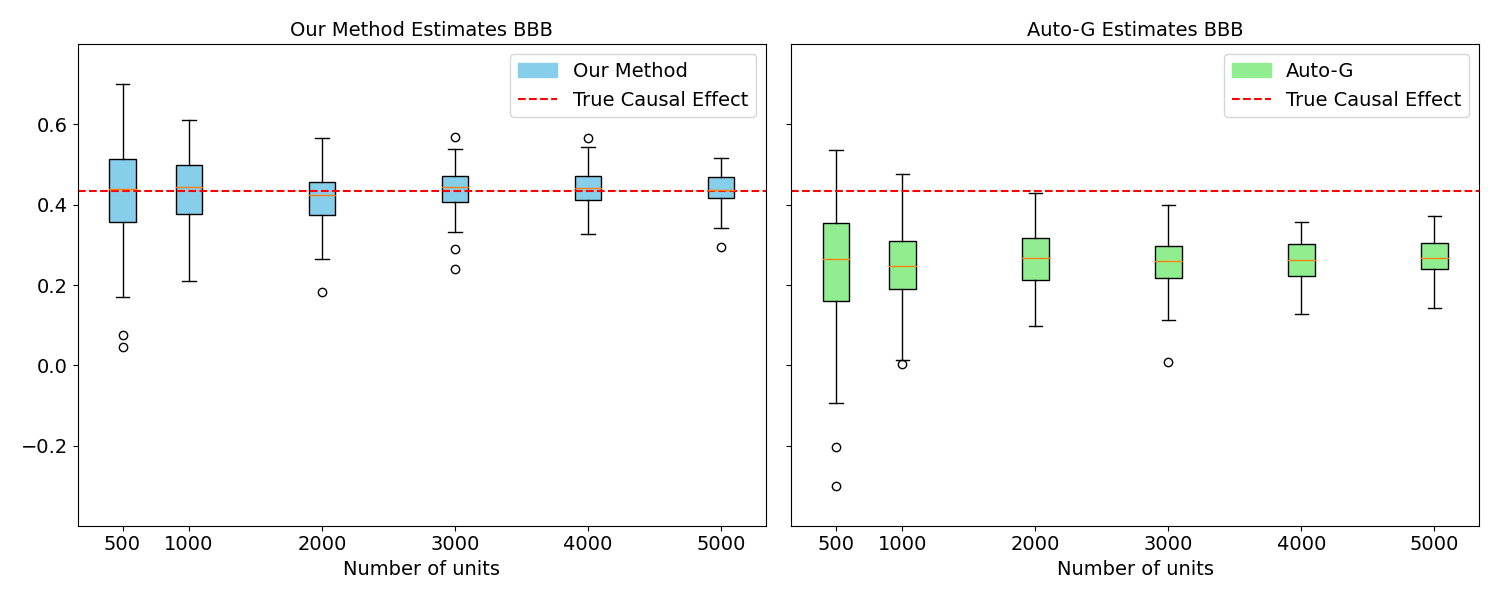}
\caption{Network effect estimation results for our method and auto-g-computation.}
\label{fig:BBB}
\end{figure*}

Figure \ref{fig:BBB} illustrates the consistency of our proposed method in the BBB case. In contrast, the auto-g-computation yields biased results due to violations of its assumptions. Experimental results for other cases are deferred to the Appendix, as they are quite similar. 

\subsection{Discussion and Conclusion}
In this work we developed hypothesis tests to differentiate between associations due to contagion and latent confounding in networks with full interference. We also extend auto-g-computation to settings where sets of variables may be dependent due to either contagion or latent confounding. We demonstrated our method's effectiveness using a mixture of synthetic and real network data.

Future work on this topic include exploring more sample efficient estimators (for example, those designed using pseudolikelihood functions \cite{besag1974spatial}) and methods that are robust to uncertainty in the friendship network $\F$. An interesting future direction would also be to explore the feasibility of mechanism testing and network effect identification and estimation in models where contagion and latent confounding co-occur in the same layer. \cite{sadeghi2016marginalization} studies separation criteria for graphs depicting such co-occurrences, however, to our knowledge, it is currently unknown whether smooth globally identifiable models for this class of graphs is possible in general. If global identification is not possible, developing sensitivity analysis methods for the assumption that contagion and latent confounding cannot co-occur would be an interesting future direction.

\acks{RB acknowledges support from the NSF CRII grant 2348287. The content of the information does not necessarily reflect the position or the policy of the Government, and no official endorsement should be inferred.}

\clearpage

\bibliography{clearbib}

\clearpage

\appendix

\section{Examples and Non-Examples of Segregated Graphs}
\label{app:examples_of_sg}

\begin{figure}[ht]
    \begin{center}
        \scalebox{0.85}{
    \begin{tikzpicture}[>=stealth, node distance=1.5cm]
        \tikzstyle{square} = [draw, minimum size=1.0mm, inner sep=3pt]
        \tikzstyle{every node} = [minimum size=1.0mm, inner sep=2pt]

        \begin{scope}[]
        \path[->, very thick]
            node[] (a) {$A$}
            node[right of=a] (b) {$B$}
            node[below of=a] (c) {$C$}
            node[below of=b] (d) {$D$}
            node[below right of=c, xshift=-0.3cm, yshift=0.3cm] (label) {(a)}

            (a) edge[->] (c)
            (b) edge[->] (d)
            (a) edge[-] (b)
            (c) edge[-] (d)
            ;
        \end{scope}
              
        \begin{scope}[xshift=3.5cm]
            \path[->, very thick]
            node[] (a) {$A$}
            node[right of=a] (b) {$B$}
            node[below of=a] (c) {$C$}
            node[below of=b] (d) {$D$}
            node[below right of=c, xshift=-0.3cm, yshift=0.3cm] (label) {(b)}

            (a) edge[->] (b)
            (a) edge[->] (c)
            (b) edge[->] (d)
            (a) edge[<->,bend left = 45] (b)
            (c) edge[-] (d)
            ;
        \end{scope}

        \begin{scope}[xshift=7cm]
        \path[->, very thick]
            node[] (a) {$A$}
            node[right of=a] (b) {$B$}
            node[below of=a] (c) {$C$}
            node[below of=b] (d) {$D$}
            node[below right of=c, xshift=-0.3cm, yshift=0.3cm] (label) {(c)}

            (a) edge[->] (c)
            (d) edge[->] (b)
            (a) edge[-] (b)
            (c) edge[-] (d)
            ;
        \end{scope}

       \begin{scope}[xshift=10.5cm]
            \path[->, very thick]
            node[] (a) {$A$}
            node[right of=a] (b) {$B$}
            node[below of=a] (c) {$C$}
            node[below of=b] (d) {$D$}
            node[below right of=c, xshift=-0.3cm, yshift=0.3cm] (label) {(d)}

            (a) edge[-] (b)
            (a) edge[->] (c)
            (b) edge[->] (d)
            (a) edge[<->,bend left = 45] (b)
            (c) edge[-] (d)
            ;
        \end{scope}

        \begin{scope}[xshift=14cm]
            \path[->, very thick]
            node[] (a) {$A$}
            node[right of=a] (b) {$B$}
            node[below of=a] (c) {$C$}
            node[below of=b] (d) {$D$}
            node[below right of=c, xshift=-0.3cm, yshift=0.3cm] (label) {(e)}

            (a) edge[-] (b)
            (a) edge[<->] (c)
            (b) edge[<->] (d)
            (c) edge[-] (d)
            ;
        \end{scope}
        
    \end{tikzpicture}
        }
    \vspace{-10pt}
    \end{center}
    \caption{The graphs shown in (a) and (b) are examples of segregated graphs. The graphs shown in (c), (d), and (e) are not segregated graphs.}
    \label{fig:examples_non_examples_of_sg}
\end{figure}

The two graphs in Figures~\ref{fig:examples_non_examples_of_sg}(a) and (b) are indeed SGs as they satisfy properties (i-iii) listed in the definition of SGs in Section~\ref{sec:setup}. The graph in Figure~\ref{fig:examples_non_examples_of_sg}(c) is not an SG due to the existence of a partially directed cycle, violating property (iii). The graph in Figure~\ref{fig:examples_non_examples_of_sg}(d) is not an SG due to $A$ and $B$ being connected by both a bidirected and undirected edge, violating property (i). Finally, Figure~\ref{fig:examples_non_examples_of_sg}(e) is not an SG as there exist vertices that serve as the endpoints of both a bidirected and undirected edge, violating property (ii).

\section{Detailed Description of s-separation}
\label{app:s-separation}

To describe s-separation in an SG $\G(V)$ with vertices $V$, we require the following graphical definitions. For brevity, we sometimes use $\circcirc$ to denote situations where any of the three types of edges is valid and $\circright$ to indicate that only a directed or bidirected edge is valid.

\subsection{Graph terminology}

A \emph{walk} in $\G(V)$ is an alternating sequence of vertices and edges $V_1 \circcirc V_2  \cdots \circcirc V_K$, where every $V_k \circcirc V_{k+1}$ is an edge present in $\G$. A \emph{path} is a walk consisting of unique vertices and edges. A \emph{section} of a walk is defined as a maximal subwalk that only consists of undirected edges (by definition a section could consist of just a single vertex). Two vertices $V_i$ and $V_j$ are said to be connected by a \emph{collider section} if there exists a walk between $V_i$ and $V_j$ of the form $V_i \circright \sigma \circleft V_j$, where $\sigma$ is a section. A \emph{partially directed walk} between vertices $V_i$ and $V_j$ is a walk starting at $V_i$ and ending at $V_j$ where every edge is either an undirected edge or a directed edge pointing from $V_{k}$ to $V_{k+1}$.  The \emph{anterior} of a set of variables $S$ denoted as $\ant_\G(S)$, is the set $S$ as well as any vertices in $\G$ that have a partially directed walk to any vertex in $S$. We use $\G_S$ to denote a \emph{subgraph} of $\G(V)$ consisting of only vertices in $S$ and the edges between them. Finally, given a subgraph $\G_S$ we use $\G_S^a$ to denote an undirected graph---sometimes called an \emph{augmented graph}---containing vertices in $S$ and edges $S_i - S_j$ if any edge $S_i \circcirc S_j$ exists in $\G_S$ or $S_i$ and $S_j$ are connected by a walk in $\G_S$ consisting of just collider sections.\footnote{Two examples of such walks are $S_i \leftrightarrow A \leftrightarrow B \leftrightarrow S_j$ and $S_i \rightarrow A - B - C \leftarrow S_j.$}

Finally, given disjoint sets of vertices $X, Y, Z$ such that $(X \cup Y \cup Z) \subseteq V$, the sets $X$ and $Y$ are said to be \emph{s-separated} given $Z$ in an SG $\G(V)$ if at least one vertex in $Z$ is present in each path between $X$ and $Y$ in $\G_{\ant_\G(X \cup Y \cup Z)}^a$. In the main text, we described the relation between s-separation and the global Markov property of SGs. In our proofs, we also occasionally invoke the following standard result in undirected graphs (these undirected graphs arise as a consequence of augmentation when checking s-separation queries). The Markov blanket of a vertex $V_i$ in the undirected graph $\G(V)$, denoted $\mb_\G(V_i)$, is defined as the set of all vertices $V_j$ that share an undirected edge with $V_i$. In an undirected graph $\G(V)$, we have $V_i \ci V \setminus \mb_\G(V_i) \cup \{V_i\} \mid \mb_\G(V_i)$ in $p(V)$.

\section{Proof of Lemma~\ref{lemma:cis}}
\label{app:proof_lemma_cis}
\begin{proof}
We first prove \eqref{eq:y-ci} holds under the null hypothesis, i.e., edges in the $Y$ layer are all undirected.
Let $\ant_\G(\cdot)$ denote the anterior of the vertices in the independence query \eqref{eq:y-ci}. Under the null, (regardless of connections in other layers) there are no walks from $Y_i$ to any other vertex in $\ant_\G(\cdot)$ that consist of just collider sections, since all walks starting at $Y_i$ must start with either $Y_i \leftarrow \cdots$ or $Y_i - \cdots$ which leads to at least one non-collider section. Based on this observation, the only edges incident to $Y_i$ in the augmented graph $\G^a_{\ant_\G(\cdot)}$ are undirected versions of those that already exist in $\G_{\ant_\G(\cdot)}$. This leads to the following edges in the augmented graph: edges $Y_i - L_k \ \forall\  L_k \in L_{\F_i^{(0, 1)}}$, edges $Y_i - A_k \ \forall\  A_k \in A_{\F_i^{(0, 1)}}$ , and edges $Y_i - Y_k\  \forall\ Y_k \in Y_{\F_i^{(1)}}$. The conditioning set in \eqref{eq:y-ci} is then a superset of vertices that have edges incident to $Y_i$. Further, no $Y_j \in Y_{\F_i^{(2)}}$ is contained in this set. The result then follows, since $Y_i$ and in $Y_{\F_i^{(2)}}$ are separated in $\G^a_{\ant_\G(\cdot)}$ given the conditioning set.

We now show that $\eqref{eq:y-ci}$ does not hold under the alternative hypothesis, i.e., edges in the $Y$ layer are all bidirected. In this case, there exist walks in $\G_{\ant_\G(\cdot)}$ from $Y_i$ to $Y_j \in Y_{\F_i^{(2)}}$  consisting of just collider sections---for example, ones of the form $Y_i \leftrightarrow Y_k \leftrightarrow Y_j$, where $Y_k \in Y_{\F^{(1)}}$. Under the assumption of  faithfulness, the result follows immediately as there now exists $Y_i - Y_j$ in  $\G^a_{\ant_\G(\cdot)}$ and the two vertices are not s-separated based on the proposed conditioning set.

The proofs for \eqref{eq:l-ci} and \eqref{eq:a-ci} are nearly identical using the appropriate $\G_{\ant_\G(\cdot)}$ and $\G^a_{\ant_\G(\cdot)}$ based on vertices provided in the respective conditional independence queries.
\end{proof}

\section{Proof of Lemma~\ref{lemma:coding}}
\label{app:proof_lemma_coding}

\begin{proof}
We first show that the factorizations in \eqref{eq:coding_y} hold under both the null and alternative hypotheses. Let $L_{{\cal CL}}, A_{{\cal CL}}, Y_{{\cal CL}}$ be variables that appear in the coding likelihoods in \eqref{eq:coding_y}, and for each $\person_i \in \I^6(\F)$ let ${\cal O}_i = L_{\F_i^{(0, 1, 2, 3)}} \cup A_{\F_i^{(0, 1, 2, 3)}} \cup Y_{\F_i^{(0, 1, 2)}}$ and ${\cal O}_{-i} = (L_{{\cal CL}} \cup A_{{\cal CL}} \cup Y_{{\cal CL}}) \setminus {\cal O}_i$. Notice for any $\person_i \in \I^6(\F)$, the variables ${\cal O}_{-i}$ belong to units at least 4 degrees of separation away from $\person_i$. If this were not the case, there would be a path between $\person_i$ and $\person_j$ in $\F$ that is shorter than $6$ edges, contradicting the antecedent that they are $6$ degrees apart. Thus, to show factorizations of the coding likelihood function, it is sufficient to show that for each $\person_i \in \I^6(\F)$ and any $\person_j$ more than 3 degrees of separation away from $\person_i$, we have that $Y_i \ci L_j, A_j, Y_j \mid \cdot$ and $Y_i \ci L_j, A_j, Y_j \mid \cdot, Y_{\F_i^{(2)}}$ under both the null and alternative hypotheses.

Under the null, we have only undirected edges in the $Y$ layer so the anterior for $Y_i$ could potentially be the entire graph. However, similar to the proof for Lemma~\ref{lemma:cis}, the vertices in $Y_{\F_i^{(1)}}\cup A_{\F_i^{(0, 1)}}\cup L_{\F_i^{(0, 1)}}$ form the Markov blanket of $Y_i$ in $\G^a_{(L \cup A \cup Y)}$. As the conditioning sets in \eqref{eq:coding_y} are supersets of the Markov blankets of each $Y_i$ and variables of $\person_j$ are not included in them, the aforementioned independences for factorization hold.

Under the alternative, we have only bidirected edges in the $Y$ layer. Thus, for some $\person_i \in \I^6(\F)$ and any $\person_j$ more than 3 degrees of separation away from $\person_i$ in $\F$, $Y_{\F_i^{(3)}}$ is not included in the following two augmented graphs: $\G^a_1 \equiv \G^a_{\ant_\G(Y_i, L_j, A_j, Y_j, \cdot)}$ and $\G^a_2 \equiv \G^a_{\ant_\G(Y_i, L_j, A_j, Y_j, \cdot, Y_{\F_i^{(2)}})}$. In $\G_1$, the variable $Y_i$ has a walk consisting of just collider sections to $L_k \in L_{\F_i^{(2)}}$ and $A_k \in A_{\F_i^{(2)}}$, which implies the existence of edges $Y_i - L_k$ and $Y_i - A_k$ in the augmented graph $\G^a_1$. Together with variables $\{V_k \mid Y_i \circleft V_k \in \G_1\}$, the neighbors of $Y_i$ in $\G^a_1$ are: $L_k \in L_{\F_i^{(0,1,2)}}$, $A_k \in A_{\F_i^{(0,1,2)}}$, and $Y_k \in Y_{\F_i^{(1)}}$. Since they are a subset of `$\cdot$' in the query $Y_i \ci L_j, A_j, Y_j \mid \cdot$, the coding likelihood ${\cal CL}(\beta_Y)$ factorizes under the alternative. Similarly in $\G_2$, $Y_i$ has a walk consisting of just collider sections to $L_k \in L_{\F_i^{(2,3)}}$, $A_k \in A_{\F_i^{(2,3)}}$, and $Y_k \in Y_{\F_i^{(2)}}$ (but not $Y_k \in Y_{\F_i^{(3)}}$ as mentioned earlier). Thus, $\mb_{\G^a_2}(Y_i) = L_{\F_i^{(0,1,2,3)}} \cup  A_{\F_i^{(0,1,2,3)}} \cup Y_{\F_i^{(1,2)}}$. These neighbors is exactly the conditioning set in $Y_i \ci L_j, A_j, Y_j \mid \cdot, Y_{\F_i^{(2)}}$, which proves the independence property needed to factorize ${\cal CL}(\gamma_Y)$ under the alternative.

The proofs for \eqref{eq:coding_l} and \eqref{eq:coding_a} are nearly identical using the appropriate independence queries based on the variables that appear in the \eqref{eq:coding_l} and \eqref{eq:coding_a}.
\end{proof}

\section{Proof of Theorem~\ref{theorem:id_formula}}
\label{app:proof_lemma_id}

\begin{proof}
Let ${\cal B}(\G)$ denote the set of maximal undirected connected components of $\G(V \cup H)$. Given a set of variables $A \subset (V \cup H)$, the post-intervention distribution where we intervene on $A$ is obtained via an extension of the g-formula \citep{pearl2009causality, robins1986new} for chain graphs \citep{lauritzen2002chain},
\begin{align*}
p(V\setminus A \mid \doo(a)) = \!\!\!\prod_{B \in {\cal B}(\G)} p(B \setminus A \mid \pa_\G(B), B \cap A) \bigg\vert_{A=a}.
\end{align*}
While these post-intervention distributions are always identified when there are no hidden variables $(H=\emptyset)$, this is not the case in general.

    Under the assumptions of Figure~\ref{fig:setup}(b), we can partition $H$ into distinct exogenous sets $H_L, H_A, H_Y$ (any of which could be empty if the respective layer contains undirected edges instead of bidirected edges) that have outgoing edges to vertices in $L, A, Y$ respectively. Then, by the CG g-formula,
    \begin{align*}
    p(L, Y &\mid \doo(a)) = \sum_{H_L, H_A, H_Y} \big(p(H_L) \times p(L \mid H_L)  \\
    &\times p(H_A) \times p(H_Y) \times p(Y\mid H_Y, A=a, L) \big).
\end{align*}
By evaluating the sum over $H_A$ and gathering like terms we can simplify the right hand side to,
\begin{align*}
    \bigg(\sum_{H_L}p(H_L, L)\bigg) \bigg(\sum_{H_Y} p(H_Y) \times p(Y \mid A=a, L)\bigg).
\end{align*}
Finally, since $H_Y \ci A, L$ by applying s-separation in $\G(V \cup H)$ we can further simplify to,
\begin{align*}
    \bigg(\sum_{H_L}p(H_L, L)\bigg) \bigg(\sum_{H_Y} p(H_Y, Y \mid A=a, L)\bigg).
\end{align*}
This gives the following identifying formula in terms of just observed variables,
\begin{align*}
    p(L, Y &\mid \doo(a)) = p(L) \times p(Y\mid A=a, L).
\end{align*}
The result follows as,
\begin{align*}
    \mathbb{E}[Y_i \mid \doo(a)]&=\sum_{L, Y} p(L, Y \mid \doo(a)) \times Y_i =\sum_{L, Y} p(L) \times p(Y \mid A=a, L) \times Y_i.
\end{align*}
\end{proof}

\section{Proof of Different Identifying Functionals for $\mathbb{E}[Y_i \mid \doo(Y_{-i}=y_{-i})]$}
\label{app:proof_lemma_different_id}

\begin{proof}
First consider the case when the edges in the $Y$ layer are undirected. In this case $H_Y = \emptyset$, while $H_L$ and $H_A$ may or may not be empty. Applying the chain graph g-formula gives us,
\begin{align*}
p(L, A, Y_i \mid \doo(y_{-i})) = \!\!\sum_{H_L, H_A} p(H_L) \times p(L \mid H_L) \times p(H_A) \times p (A\mid H_A, L) \times p(Y_i \mid y_{-i}, A, L).
\end{align*}
Since $H_A \ci L$ in $\G(V\cup H)$ we can simplify the above and gather like terms to get,
\begin{align*}
\bigg(\sum_{H_L}p(H_L, L)\bigg) \bigg(\sum_{H_A} p(H_A, A \mid L)\bigg) \times p(Y \mid y_{-i}, A, L).
\end{align*}
This gives the following identifying formula in terms of just observed variables,
\begin{align*}
p(L, A, Y_i \mid \doo(y_{-i}) = p(L) \times p(A\mid L) \times p(Y\mid y_{-i}, A, L).
\end{align*}
The result follows as,
\begin{align*}
    \mathbb{E}[Y_i \mid \doo(y_{-i})]&=\sum_{L, A, Y_i} p(L, A, Y_i \mid \doo(y_{-i})) \times Y_i =\sum_{L, A, Y_i} p(L, A) \times p(Y \mid y_{-i}, A, L) \times Y_i.
\end{align*}

Now we turn to the case when the edges in the $Y$ layer are bidirected. Applying the chain graph g-formula in this setting gives us,
{\small
\begin{align*}
p(L, A, Y_i \mid \doo(y_{-i})) = \!\!\!\!\sum_{H_L, H_A, H_Y} p(H_L) \times p(L \mid H_L) \times p(H_A) \times p (A\mid H_A, L) \times p(H_Y) \times p(Y_i \mid H_Y, A, L).
\end{align*}}
Simplifying this functional based on independences $H_A \ci L$ and $H_Y \ci A, L$ gives us,
\begin{align*}
p(L, A, Y_i \mid \doo(y_{-i})) = p(L) \times p(A\mid L) \times p(Y_i \mid A, L) = p(L, A, Y_i).
\end{align*}
The result follows as,
\begin{align*}
    \mathbb{E}[Y_i \mid \doo(y_{-i})]&=\sum_{L, A, Y_i} p(L, A, Y_i \mid \doo(y_{-i})) \times Y_i =\sum_{Y_i} p(Y_i) \times Y_i.
\end{align*}

\end{proof}

\section{Proof of Lemma~\ref{lemma:coding_ricf}}
\label{app:proof_lemma_coding_ricf}
The proof for this is very similar to the proof in Appendix~\ref{app:proof_lemma_coding} and involves checking some relatively simple s-separation conditions compared to those in Appendix~\ref{app:proof_lemma_coding} as we never condition on variables in colliders or collider sections. Thus, we omit further details of the proof.

\clearpage

\section{Gibbs Samplers for $L$ and $Y$}
\label{app:gibbs_samplers}

The Gibbs samplers for drawing samples from $p(L)$ and $p(Y \mid A, L)$ when the edges in the $L$ and $Y$ layer are undirected are described in Algorithms~\ref{alg:Gibbs_L} and~\ref{alg:Gibbs_Y}. In these algorithms, $<i$ denotes all units in $\F$ that precede unit $i$ in a total ordering of all units and $>i$ denotes all units that come after unit $i$. In our experiments we set the number of draws $M$ from the Gibbs sampler to be $0.3\times N$, the thinning interval $T$ to be $3$, and the burn-in period $m^*$ to be $200$.



\begin{algorithm}[H]
\caption{Gibbs sampler $L$}
\label{alg:Gibbs_L}
\begin{algorithmic}[1]
\State \textbf{Inputs:} SG $\G$;  Parameters  $\theta_{L_i - L_j}$
\State Initialize an empty list $L_{\text{draws}}$ of size $M$ corresponding to number of draws we want
\State Initialize a random vector $L^{(0)}$

\For{$m=0, 1, \ldots, T \times M + m^*$} \quad // $m^*$ is the burn-in period; $T$ is the thinning interval
    \For{$i=1, \ldots, N$}
    \State $L_i^{(m+1)} \gets \text{ draw from }p(L_i \mid L_{<i}^{(m+1)}, L_{>i}^{(m)}; \theta_{L_i - L_j})$
    \EndFor 
    \If{($m > m^*$) and ($m$ mod $T = 0$)}
        \State Append $L^{(m)}$ to $L_{\text{draws}}$
    \EndIf
\EndFor
\State \textbf{return} $L_{\text{draws}}$ 
\end{algorithmic}
\end{algorithm}

\begin{algorithm}[H]
\caption{Gibbs sampler $Y$}
\label{alg:Gibbs_Y}
\begin{algorithmic}[1]
\State \textbf{Inputs:} SG $\G$; Parameters for  $\theta_{Y_i - Y_j}$; Previously sampled $L_{\text{draws}}$, a size $M$ list; Treatment assignment vector $a$.
\State Initialize $Y_{\text{draws}}$, an empty list of size $M$

\For{$m=1, \ldots, M$}
    \State $L^{(m)} \gets$ the $m$-th element of $L_{\text{draws}}$
    \State Initialize a random vector $Y^{(0)}$;
    \For{$m'=0, 1 \ldots, m^*$} \quad // $m^*$ is the burn-in period.
        \For{$i=1, \ldots, N$}
        \State $Y_i^{(m'+1)} \gets \text{ draw from }p(Y_i \mid Y_{<i}^{(m'+1)}, Y_{>i}^{(m')}, A=a, L^{(m)}; \theta_{Y_i - Y_j})$
        \EndFor 
    \EndFor
    \State Append $Y^{(m^*)}$ to $Y_{\text{draws}}$ 
\EndFor

\State \textbf{return} $Y_{\text{draws}}$ 
\end{algorithmic}
\end{algorithm}

\clearpage

\section{Experiment Details}
\label{app:experiment-details}

\subsection{Data Generating Process (DGP) for Evaluating the Likelihood Ratio Tests}

In this section, we outline the DGPs for the two configurations of $\G$: one with undirected edges in all three layers and the other with bidirected edges in all layers; and directed edges in both configurations follow the set up in the partially determined SG $\G_{\dyad}$ in Figure \ref{fig:setup} (b).

For the configuration where all layers contain undirected edges, we generate $L$, $A$, and $Y$ as vectors of binary values using the following Gibbs factors:

\begin{align*}
   L_i &\sim \text{Bernoulli}(\text{expit}(-0.1 * \sum_{j \in \F_i^{(1)}}L_j)); \\
   A_i &\sim \text{Bernoulli}(\text{expit}(0.8 * L_i - 0.1 * \sum_{j \in \F_i^{(1)}}L_j - 0.1 * \sum_{j \in \F_i^{(1)}}A_j )); \\
   Y_i &\sim \text{Bernoulli}(\text{expit}(0.8 * L_i + 1.7 * A_i - 0.1 * \sum_{j \in \F_i^{(1)}}L_j - 0.1 * \sum_{j \in \F_i^{(1)}}A_j - 0.1 * \sum_{j \in \F_i^{(1)}}Y_j)).
\end{align*}

Using these Gibbs factors, we generate values layers by layers in topological order. First, we perform Gibbs sampling to generate $L$. Next, we fix the $L$ vector and use it as inputs to generate $A$. Finally, $L$ and $A$ are used as inputs to generate $Y$. For the Gibbs sampling process at each layer, we use a burn-in period equaling 200 multiplied by the number of units in the network $\F$.

For the configuration where all layers contain bidirected edges, we generate $L$, $A$, and $Y$ as vectors of binary values by first explicitly generating the hidden variables $H_{ij}^{L}$, $H_{ij}^{A}$, and $H_{ij}^{Y}$ corresponding to the bidirected edges $L_i \biedge L_j$, $A_i \biedge A_j$, and $Y_i \biedge Y_j$, respectively, for each connected pair $(i, j)$ in $\F$. After $L$, $A$, and $Y$ are generated, we discard the $H$ values to make them unobserved. The parameters of this DGP are as follows:

\begin{align*}
   &H_{ij}^{L}, H_{ij}^{A}, H_{ij}^{Y} \sim \text{Normal}(0, 1); \\
   &L_i \sim \text{Bernoulli}(\text{expit}(5 * \sum_{k \in \F_i^{(1)}} H_{ik}^L)); \\ 
   &A_i \sim \text{Bernoulli}(\text{expit}(0.2 * L_i + 0.1 * \sum_{k \in \F_i^{(1)}}L_k + 5 * \sum_{k \in \F_i^{(1)}} H_{ik}^A)); \\
   &Y_i \sim \text{Bernoulli}(\text{expit}(0.2 * L_i - 0.3 * A_i + 0.1 * \sum_{k \in \F_i^{(1)}}L_k - 0.2 * \sum_{k \in \F_i^{(1)}}A_k + 5 * \sum_{k \in \F_i^{(1)}} H_{ik}^Y));
\end{align*}

\subsection{Parametric Models for Evaluating the Likelihood Ratio Tests}

When conducting the likelihood ratio tests, we use the following nested parametric models. Models \eqref{eq:null_l_appendix} and \eqref{eq:alt_l_appendix} correspond to the null and alternative models we fit when conducting the likelihood ratio test for the $L$ layer. Similarly, \eqref{eq:null_a_appendix} and \eqref{eq:alt_a_appendix} are the null and alternative models for the $A$ layer, and \eqref{eq:null_y_appendix} and \eqref{eq:alt_y_appendix} are for the $Y$ layer.

\begin{align}
&p(L_i=1 \mid L_{\F_i^{(1)}}; \beta) = \text{expit}(\beta_0 + \beta_1 \sum_{k \in {\F_i^{(1)}}}L_k)\label{eq:null_l_appendix} \\
    &p(L_i=1 \mid L_{\F_i^{(1,2)}}; \gamma) = \text{expit}(\gamma_0 + \gamma_1 \sum_{k \in {\F_i^{(1)}}}L_k + \gamma_2 \sum_{k \in {\F_i^{(1)}}}L_k)\label{eq:alt_l_appendix} \\ \nonumber \\ \nonumber \\ &p(A_i=1 \mid A_{\F_i^{(1)}},  L_{\F_i^{(0,1,2,3)}}; \beta) = \text{expit} (\label{eq:null_a_appendix}\beta_0 + \beta_1 \sum_{k \in {\F_i^{(1)}}}A_k + \\ \nonumber  & \beta_2 L_i + \beta_3 \sum_{k \in {\F_i^{(1)}}}L_k + \beta_4 \sum_{k \in {\F_i^{(2)}}}L_k + \beta_5 \sum_{k \in {\F_i^{(3)}}}L_k) \\ &p(A_i=1 \mid A_{\F_i^{(1,2)}},  L_{\F_i^{(0,1,2,3)}}; \gamma) = \text{expit} (\label{eq:alt_a_appendix}\gamma_0 + \gamma_1 \sum_{k \in {\F_i^{(1)}}}A_k + \gamma_2 \sum_{k \in {\F_i^{(2)}}}A_k +\\ \nonumber & \gamma_3 L_i + \gamma_4 \sum_{k \in {\F_i^{(1)}}}L_k + \gamma_5 \sum_{k \in {\F_i^{(2)}}}L_k + \gamma_6 \sum_{k \in {\F_i^{(3)}}}L_k) \\ \nonumber \\ \nonumber \\ &p(Y_i=1 \mid Y_{\F_i^{(1)}}, A_{\F_i^{(0,1,2,3)}}, L_{\F_i^{(0,1,2,3)}}; \beta) = \text{expit} (\label{eq:null_y_appendix}\beta_0 + \beta_1 \sum_{k \in {\F_i^{(1)}}}Y_k + \\ & \nonumber \beta_2A_i + \beta_3 \sum_{k \in {\F_i^{(1)}}}A_k + \beta_4 \sum_{k \in {\F_i^{(2)}}}A_k + \beta_5 \sum_{k \in {\F_i^{(3)}}}A_k + \beta_6 L_i + \beta_7 \sum_{k \in {\F_i^{(1)}}}L_k + \\ \nonumber & \beta_8 \sum_{k \in {\F_i^{(2)}}}L_k + \beta_9 \sum_{k \in {\F_i^{(3)}}}L_k) \\
     &p(Y_i=1 \mid Y_{\F_i^{(1,2)}}, A_{\F_i^{(0,1,2,3)}},  L_{\F_i^{(0,1,2,3)}}; \gamma) = \text{expit} (\label{eq:alt_y_appendix}\gamma_0 + \gamma_1 \sum_{k \in {\F_i^{(1)}}}Y_k + \gamma_2 \sum_{k \in {\F_i^{(2)}}}Y_k + \\&\nonumber \gamma_3A_i + \gamma_4 \sum_{k \in {\F_i^{(1)}}}A_k + \gamma_5 \sum_{k \in {\F_i^{(2)}}}A_k + \gamma_6 \sum_{k \in {\F_i^{(3)}}}A_k + \gamma_7 L_i + \gamma_8 \sum_{k \in {\F_i^{(1)}}}L_k + \\& \nonumber \gamma_9 \sum_{k \in {\F_i^{(2)}}}L_k + \gamma_{10} \sum_{k \in {\F_i^{(3)}}}L_k) 
\end{align}

\subsection{DGPs for Evaluating the Causal Effect Estimation Method}

We now outline the DGPs for $L$, $A$, and $Y$ in this experiment. Recall that $N$ denotes the number of units in $\F$. For layers with undirected edges, we use Gibbs sampling with a burn-in period of $m^*=200$. We keep the sample immediately after the burn-in period as the sampled data. When deriving the ground-truth expectations $\mathbb{E}[Y \mid \text{do}(A = \textbf{1})]$ and $\mathbb{E}[Y \mid \text{do}(A = \textbf{0})]$, we continue running the Gibbs sampler to obtain $M=0.3\times N$ draws with a thinning interval of $T=3$ to reduce auto correlation. Thus, we average over $0.3 \times N$ samples of $Y$ to derive the ground-truths.

When the $L$ layer is connected by undirected edges, the distribution of $L_i$ for all $i \in \F$ is given by: 
\begin{align*}
  L_i &\sim \text{Bernoulli}(\text{expit}( - 0.3 + 0.4 * \sum_{k \in \F_i^{(1)}}L_k)).
\end{align*}
When the $L$ layer is connected by bidirected edges, the distribution of the entire $L$ vector is given by: 
\begin{align*}
  L &\sim MVN(\mu, \Sigma), 
\end{align*}
where $\mu$ is a vector with all entries equal to 0.7, and $\Sigma$ is a matrix with diagonal elements equal to 3.5. For an off-diagonal location $(x, y)$ in $\Sigma$, the value is 0.2 if $\person_x$ and $\person_y$ are neighbors in $\F$ and is 0 otherwise. When the $A$ layer is connected by undirected edges, the distribution of $A_i$ for all $i \in \F$ is given by:
\begin{align*}
  A_i &\sim \text{Bernoulli}(\text{expit}( 5 + 4 * L_i - 1.2 * \sum_{k \in \F_i^{(1)}}L_k -2 * \sum_{k \in \F_i^{(1)}}A_k )).
\end{align*}
When the $A$ layer is connected by bidirected edges, we generate $A$ by first explicitly sampling hidden variables $H^A_{ij}$ for each connected pair $(i, j)$ in $\F$ and then determining each entry of $A$ using the following distributions (the hidden variables are discarded after $A$ is generated):
\begin{align*}
   H_{ij}^{A} &\sim \text{Normal}(2, 1); \\
   A_i &\sim \text{Bernoulli}(\text{expit}(1.3 + 0.2 * \sum_{k \in \F_i^{(1)}} H_{ik}^A -0.4 * L_i -0.7 * \sum_{k \in \F_i^{(1)}}L_k)). 
\end{align*}
When the $Y$ layer is connected by undirected edges, the distribution of $Y_i$ for all $i \in \F$ is given by:
\begin{align*}
  Y_i &\sim \text{Bernoulli}(\text{expit}( 2 + L_i + 1.5 * A_i - 5.3 * \sum_{k \in \F_i^{(1)}}L_k + \sum_{k \in \F_i^{(1)}}A_k - 4 * \sum_{k \in \F_i^{(1)}}Y_k)).
\end{align*}
When the $Y$ layer is connected by bidirected edges, we first generate hidden variables $H^Y_{ij}$ for each connected pair $(i, j)$ in $\F$ and then generate $Y$ accordingly, using the following distributions:
\begin{align*}
   H_{ij}^{Y} &\sim \text{Normal}(0, 1); \\
   Y_i &\sim \text{Bernoulli}(\text{expit}(-1 + 2* \sum_{k \in \F_i^{(1)}} H_{ik}^Y +0.1 * L_i + A_i -0.3 * \sum_{k \in \F_i^{(1)}}L_k + \sum_{k \in \F_i^{(1)}}A_k)). \\
\end{align*}

\clearpage

\subsection{Additional Experimental Results}

In the main text, we have shown that our method produces consistent and unbiased causal estimates when all layers are connected by bidirected edges. In this section, we will show that our method is also consistent and unbiased for the other 6 cases not yet discussed in the main text---UBB, BUB, BBU, BUU, and UUB.  

\begin{figure}[h]
\centering
\includegraphics[width=0.7\linewidth]{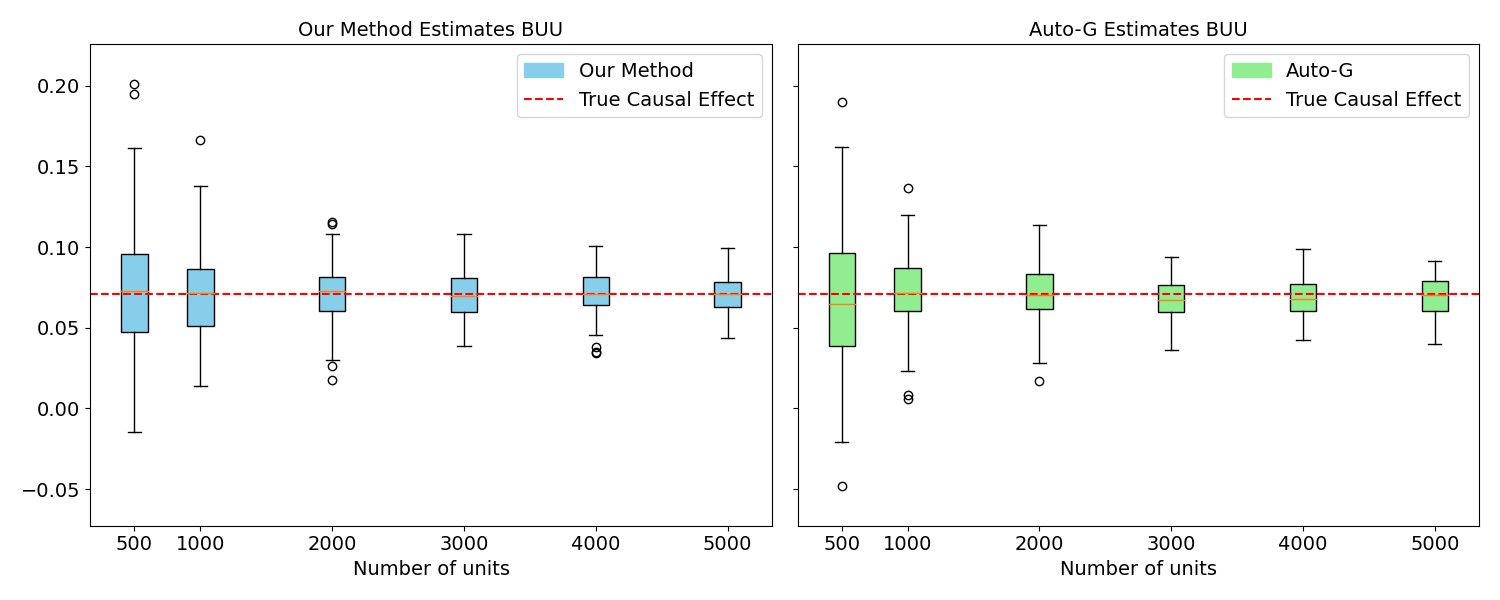}
\caption{Network effect estimation results for the BUU setup.}
\label{fig:BUU}
\end{figure}

\begin{figure}[h]
\centering
\includegraphics[width=0.7\linewidth]{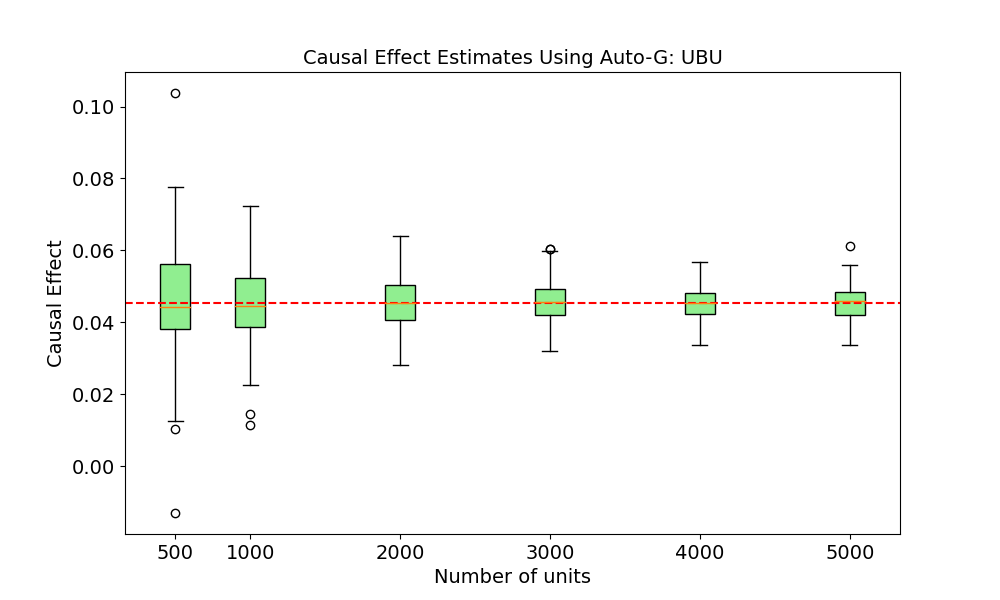}
\caption{Consistency the auto-g-computation in the UBU setup.}
\label{fig:UBU_autog}
\end{figure}

Figure \ref{fig:UBU_autog} only has a single plot because causal effects in the UBU setup can be estimated directly using the auto-g method, as discussed in the main text. For the other plots below, we see that our methods are consistent and unbiased, as the distribution of the estimates becomes narrower as the sample size increases, and the distributions all center around the ground truth. However, naive application of the auto-g-computation method to scenarios where layer $L$ or $Y$ is connected by latent confounding leads to biased results. We also find that the auto-g method has some robustness against model misspecification, as its estimates in Figure \ref{fig:BUU} are centered around the ground truth. We believe this robustness is likely due to the particular DGP we have chosen, since we have demonstrated in the main text that it is theoretically incorrect to apply the auto-g-computation method when bidirected edges exist in the $L$ layer.

\begin{figure}[h]
\centering
\includegraphics[width=0.7\linewidth]{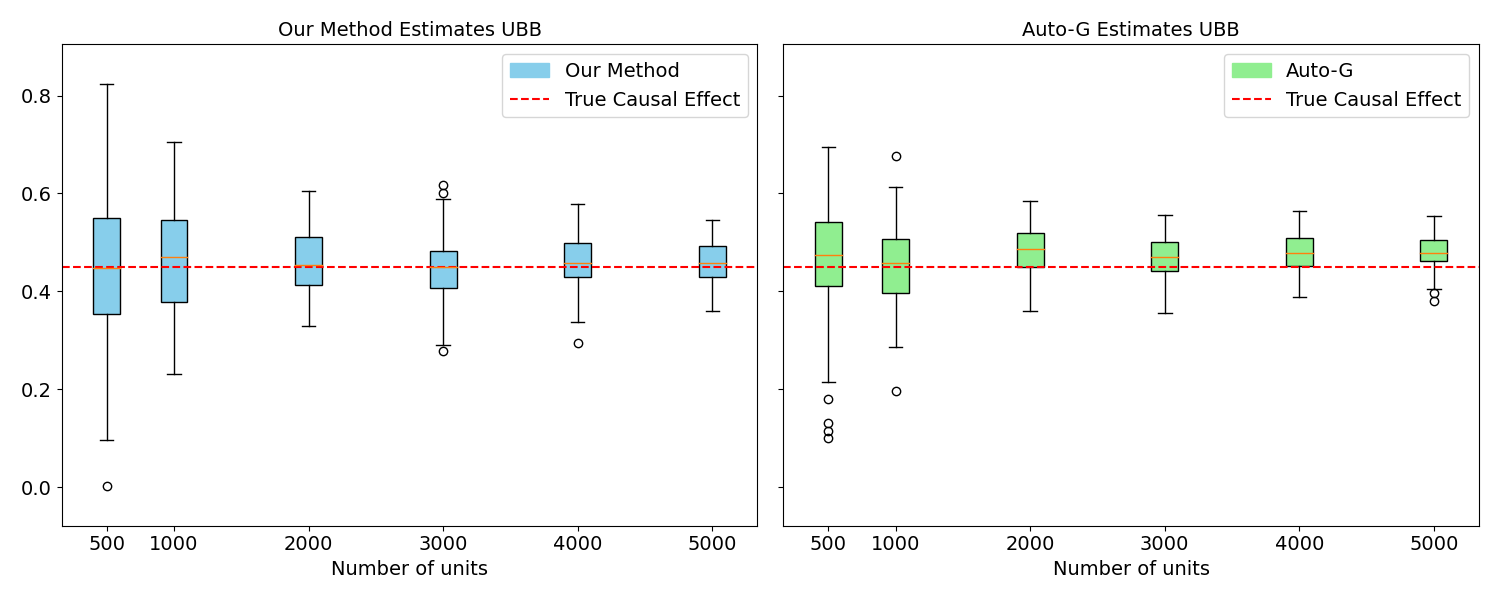}
\caption{Network effect estimation results for the UBB setup.}
\label{fig:UBB}
\end{figure}

\begin{figure}[h]
\centering
\includegraphics[width=0.7\linewidth]{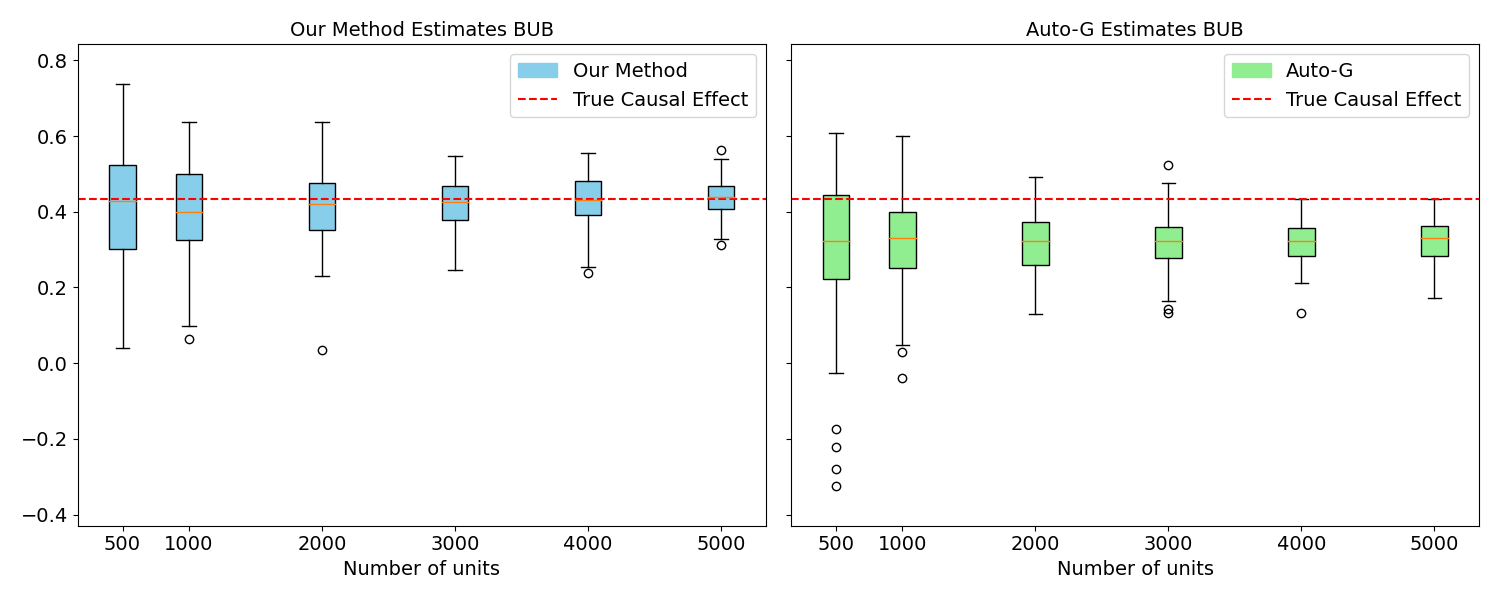}
\caption{Network effect estimation results for the BUB setup.}
\label{fig:BUB}
\end{figure}

\begin{figure}[h]
\centering
\includegraphics[width=0.7\linewidth]{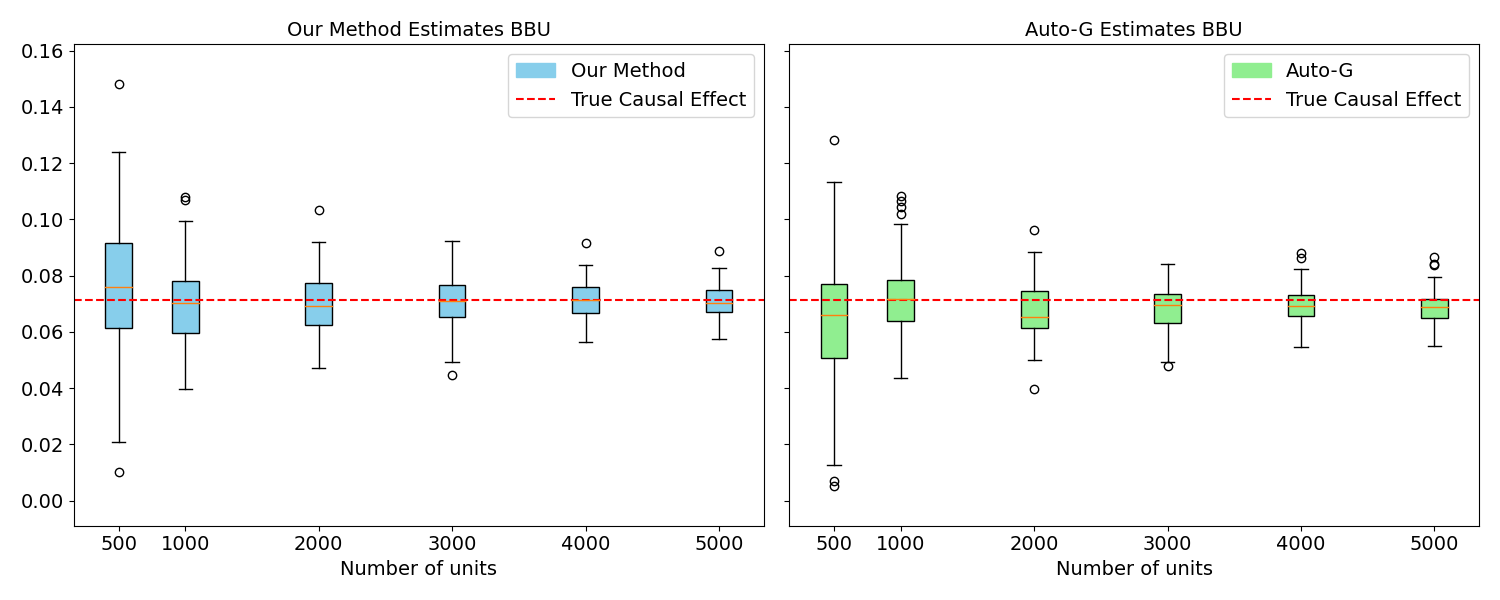}
\caption{Network effect estimation results for the BBU setup.}
\label{fig:BBU}
\end{figure}

\begin{figure}[h]
\centering
\includegraphics[width=0.7\linewidth]{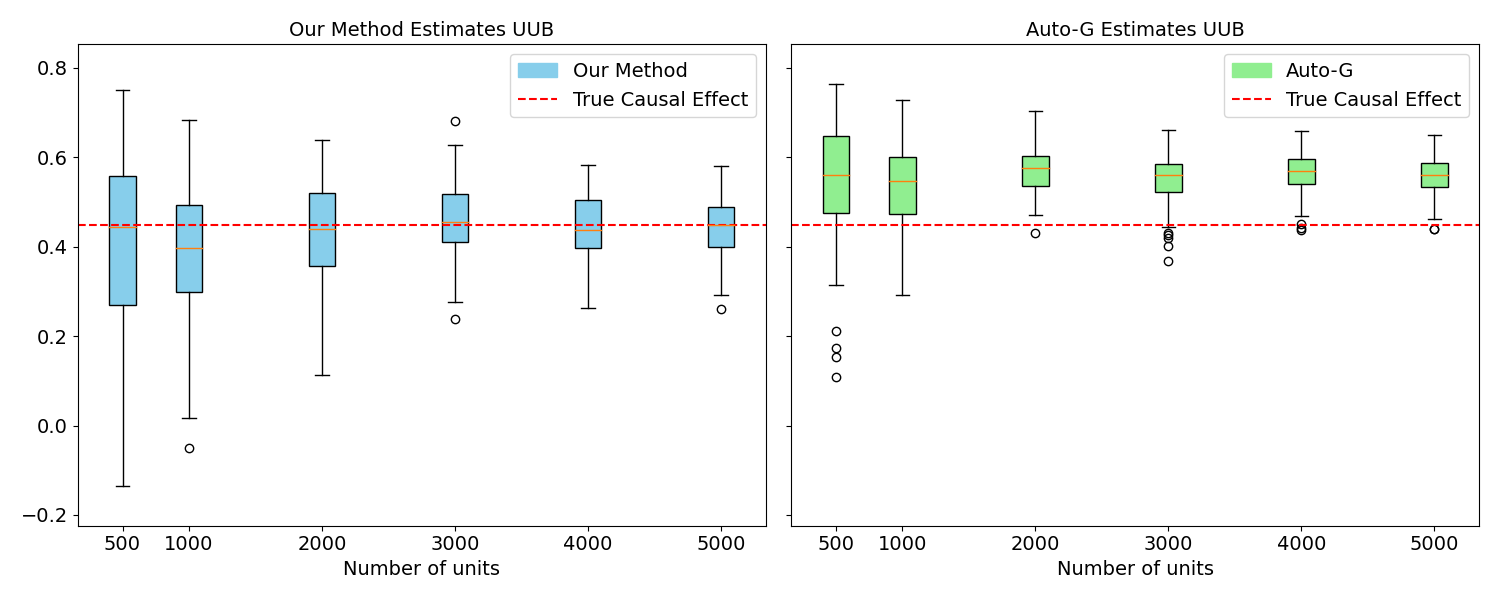}
\caption{Network effect estimation results for the UUB setup.}
\label{fig:UUB}
\end{figure}

\end{document}